\documentclass[10pt,journal,compsoc]{IEEEtran}
\ifCLASSOPTIONcompsoc
  \usepackage[nocompress]{cite}
\else
  \usepackage{cite}
\fi
\ifCLASSINFOpdf
\else
\fi

\usepackage{graphicx,multirow,epstopdf,subfigure,amssymb,amsmath}
\usepackage{array}

\begin{document}
%
\title{Facial Action Unit Detection Using Attention and Relation Learning}
%
%
%
%

\author{Zhiwen~Shao,
        Zhilei~Liu,
        Jianfei~Cai,~\IEEEmembership{Senior~Member,~IEEE,}
        Yunsheng~Wu,
        and~Lizhuang~Ma
\IEEEcompsocitemizethanks{
\IEEEcompsocthanksitem Corresponding author: Zhilei~Liu and Lizhuang~Ma.
\IEEEcompsocthanksitem Z. Shao is with the Department of Computer Science and Engineering, Shanghai Jiao Tong University, Shanghai 200240, China, and also with the MoE Key Lab of Artificial Intelligence, Shanghai Jiao Tong University, Shanghai 200240, China. E-mail: shaozhiwen@sjtu.edu.cn.
\IEEEcompsocthanksitem Z. Liu is with the College of Intelligence and Computing, Tianjin University, Tianjin 300072, China. E-mail: zhileiliu@tju.edu.cn.
\IEEEcompsocthanksitem J. Cai is with the School of Computer Science and Engineering, Nanyang Technological University, Singapore 639798. E-mail: asjfcai@ntu.edu.sg.
\IEEEcompsocthanksitem Y. Wu is with the YouTu Lab, Tencent Inc., Shanghai 200233, China. E-mail: simonwu@tencent.com.
\IEEEcompsocthanksitem L. Ma is with the Department of Computer Science and Engineering, Shanghai Jiao Tong University, Shanghai 200240, China, also with the MoE Key Lab of Artificial Intelligence, Shanghai Jiao Tong University, Shanghai 200240, China, and also with the School of Computer Science and Software Engineering, East China Normal University, Shanghai 200062, China. E-mail: ma-lz@cs.sjtu.edu.cn.
}
\thanks{Manuscript received February, 2019.}}

%
%

\markboth{IEEE Transactions on Affective Computing,~Vol.~X,~NO.~X,~X}%
{Shell \MakeLowercase{\textit{et al.}}: Bare Demo of IEEEtran.cls for Computer Society Journals}
%



\IEEEtitleabstractindextext{%
\begin{abstract}
Attention mechanism has recently attracted increasing attentions in the field of facial action unit (AU) detection. By finding the region of interest of each AU with the attention mechanism, AU-related local features can be captured. Most of the existing attention based AU detection works use prior knowledge to predefine fixed attentions or refine the predefined attentions within a small range, which limits their capacity to model various AUs. In this paper, we propose an end-to-end deep learning based attention and relation learning framework for AU detection with only AU labels, which has not been explored before. In particular, multi-scale features shared by each AU are learned firstly, and then both channel-wise and spatial attentions are adaptively learned to select and extract AU-related local features. Moreover, pixel-level relations for AUs are further captured to refine spatial attentions so as to extract more relevant local features. Without changing the network architecture, our framework can be easily extended for AU intensity estimation. Extensive experiments show that our framework (i) soundly outperforms the state-of-the-art methods for both AU detection and AU intensity estimation on the challenging BP4D, DISFA, FERA 2015 and BP4D+ benchmarks, (ii) can adaptively capture the correlated regions of each AU, and (iii) also works well under severe occlusions and large poses.
\end{abstract}

\begin{IEEEkeywords}
Channel-wise and spatial attention learning, pixel-level relation learning, facial AU detection
\end{IEEEkeywords}}

\maketitle

\IEEEdisplaynontitleabstractindextext

%
\IEEEpeerreviewmaketitle

\IEEEraisesectionheading{\section{Introduction}\label{sec:introduction}}

\IEEEPARstart{F}{acial} action units (AUs) are basic facial movements in local facial regions defined by Facial Action Coding System (FACS)~\cite{ekman1997face}, which describe fine-grained changes in facial expressions. Facial AU detection refers to determining the occurrences of different AUs in a given face image, an extension of which is to estimate AU intensities, namely facial AU intensity estimation. AU detection is an important face analysis task, which is applied in various areas such as health and entertainment by measuring human emotions. On the other hand, attention mechanism has been adopted in many structural prediction tasks such as saliency detection~\cite{kuen2016recurrent,liu2018picanet}, object recognition~\cite{xiao2015application,cao2015look}, and image captioning~\cite{you2016image,pedersoli2017areas}, where great success has been achieved. It is natural to apply the attention mechanism to find the region of interest (ROI) of each AU so that more relevant local features can be captured. However, in literature there are only a few attention based AU detection methods, in which prior knowledge is required to predefine the attentions of AUs.

\begin{figure}
\centering\includegraphics[width=0.98\linewidth]{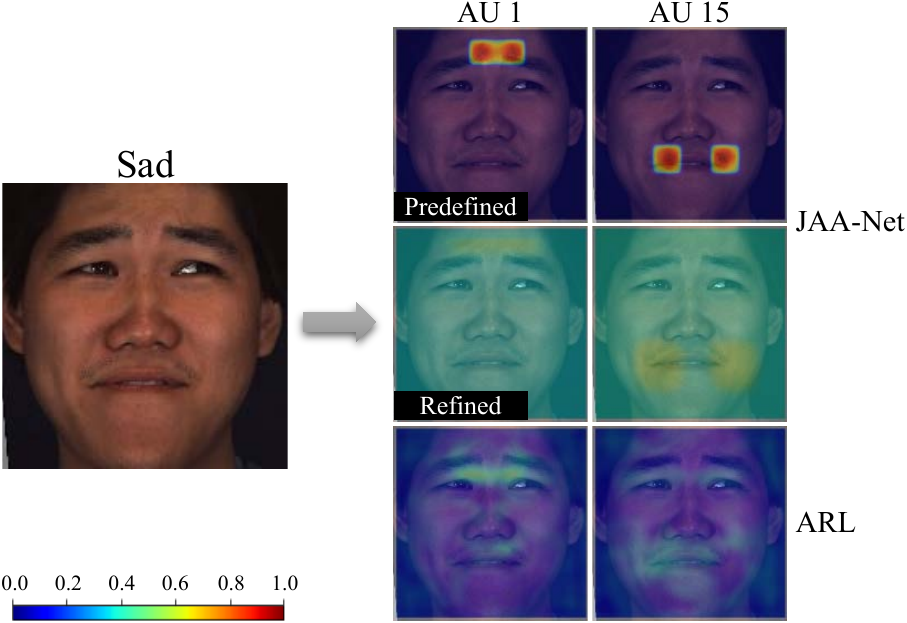}
\caption{Predefined and refined attentions of JAA-Net~\cite{shao2018deep} and learned attentions of our method ARL for AU 1 and AU 15 of an example sad face: 1) Both the predefined and refined attentions of JAA-Net for AU 1 are not highlighted in other \textit{correlated regions} including the ROI of AU 15, and vice versa. 2) The attentions of our ARL capture the correlated regions. Attention weights are visualized with the colors defined in the color bar, which are overlaid on the face image.}
\label{fig:sad_example}
\end{figure}

Since facial landmarks can provide rough locations of AUs, they have been used for predefining the AU attentions. Li et al.~\cite{li2018eac,li2017action} employed predefined attentions for AU detection, in which landmarks are used to generate the ROI of each AU with a fixed size and a fixed attention distribution. 
These methods design fixed attentions based on the prior knowledge about the location relationship between AUs and landmarks, which cannot adapt to various AUs with different sizes and non-rigid transformations. Motivated by the above limitation, Shao et al.~\cite{shao2018deep} proposed a deep learning based joint AU detection and face alignment framework called JAA-Net with an adaptive attention learning module, which is a pioneering work of learning the AU attentions adaptively. However, the refined attentions are limited in the neighboring regions of initially predefined attentions, in which correlated regions beyond the predefined attentions are not highlighted. For example, the inner brows raise (AU 1) and the lip corners depress (AU 15) simultaneously in a sad face, as shown in Fig.~\ref{fig:sad_example}, which indicates the relation between AU 1 and AU 15~\cite{pantic2000expert}. Nevertheless, both the predefined and refined attentions of a certain AU learned by JAA-Net are not highlighted in the ROIs of correlated AUs, which fail to capture the relations among AUs.

Inspired by the intuitive AU relations, many works learned to model the AU-level relations, including local pairwise AU dependencies~\cite{tong2008learning,li2013simultaneous}, global AU dependencies~\cite{wu2016constrained,zhang2016task}, and both of them~\cite{wang2013capturing,eleftheriadis2015multi}. However, these methods did not integrate the AU relation learning with the attention mechanism. Moreover, the learned AU-level relations can only provide rough correlated locations to refine the predefined attention for each AU, which is difficult to capture subtle facial appearance changes. To ensure the attention of each AU captures all the correlated regions including small regions with tiny muscle actions, pixel dependencies should be utilized to refine pixel-wise attentions. On the other hand, current attention based AU detection works~\cite{li2018eac,li2017action,shao2018deep} only consider spatial attentions, without exploiting the channel-wise attention mechanism. Since a channel-wise feature map is the result of a certain filter, channel-wise attentions can be used to select features with AU-related attributes.

In this paper, we propose a novel deep learning based Attention and Relation Learning (ARL) framework for AU detection, in which only AU labels are used to adaptively learn the implicit attentions and relations. Our framework does not rely on any other information such as the central location and boundary of each AU defined by the landmarks~\cite{li2018eac,li2017action,shao2018deep}, and thus is free from the restrictions of predefined attentions. In particular, multi-scale features shared by each AU are learned firstly, and then both channel-wise attention learning and spatial attention learning are utilized to select and extract AU-related local features. Moreover, we propose a pixel-level relation learning method with the fully-connected Conditional Random Field (CRF)~\cite{krahenbuhl2011efficient,zheng2015conditional} to refine spatial pixel-wise attentions of each AU so as to extract more relevant local features, in which the attention results of our ARL are illustrated in Fig.~\ref{fig:sad_example}. The entire framework is end-to-end for joint learning of attention and relation, without any post-processing step. Our framework can also be easily extended for AU intensity estimation without changing the network architecture.

To summarize, the main contributions of this paper are threefold:
\begin{itemize}
    \item An attention learning method is proposed for AU detection, in which both channel-wise and spatial attentions are adaptively learned using only AU labels.
    \item A pixel-level relation learning method is proposed to refine spatial pixel-wise attentions so as to capture all the correlated regions and features for each AU. To our knowledge, this is the first work of introducing joint learning of attention and pixel-level relation for AU detection.
    \item Extensive experiments demonstrate that our framework soundly outperforms the state-of-the-art approaches for both AU detection and AU intensity estimation, can adaptively capture the correlated regions of each AU, and also works well under severe occlusions and large poses.
\end{itemize}

\section{Related Work}

We review previous works that are most relevant to our method, including attention based AU detection and relation based AU detection.

\subsection{Attention Based AU Detection}

There are only a few recent works exploiting the spatial attention mechanism for AU detection. Since AUs have no distinct contour and texture and may change across persons and facial expressions, it is hard to annotate their attention labels manually. Without the ground-truth attention labels for training data, current works use the prior knowledge to predefine the AU attentions. Li et al.~\cite{li2018eac,li2017action} proposed an Enhancing and Cropping Net (EAC-Net) for AU detection by using predefined attentions to enhance and crop the ROIs of AUs. The predefined ROI of each AU has a fixed size and a fixed attention distribution, whose location is given by facial landmarks. Based on the structure of Enhancing Net (E-Net)~\cite{li2018eac}, Zhang et al.~\cite{zhang2018identity} proposed a method of adversarial training between AU detection and identity recognition. The identity classifier is trained to minimize the identity recognition loss while the feature layers are trained to maximize the identity recognition loss, so that the learned features are effective for AU detection while invariant to subject identities. Sanchez et al.~\cite{sanchez2018joint} adopted an hourglass network~\cite{newell2016stacked} for AU intensity estimation through regressing from an input image to AU attention maps. Similarly, the ground-truth attention map of each AU is predefined as a Gaussian distribution, where a certain landmark determines its central location and the AU intensity determines its amplitude and size. Instead of using predefined attentions, Shao et al.~\cite{shao2018deep} proposed the JAA-Net with an adaptive attention learning module to adaptively refine the initially predefined attention of each AU. The refined attentions are only limited in the neighboring regions of the predefined attentions, and correlated regions beyond the predefined attentions are not highlighted.

All these methods demonstrate the effectiveness of the spatial attention mechanism for AU detection. However, they all use predefined attentions or refine the attentions within a small range, which limits their capacity to model various AUs with different sizes and non-rigid transformations.

\begin{figure*}
\centering\includegraphics[width=0.98\linewidth]{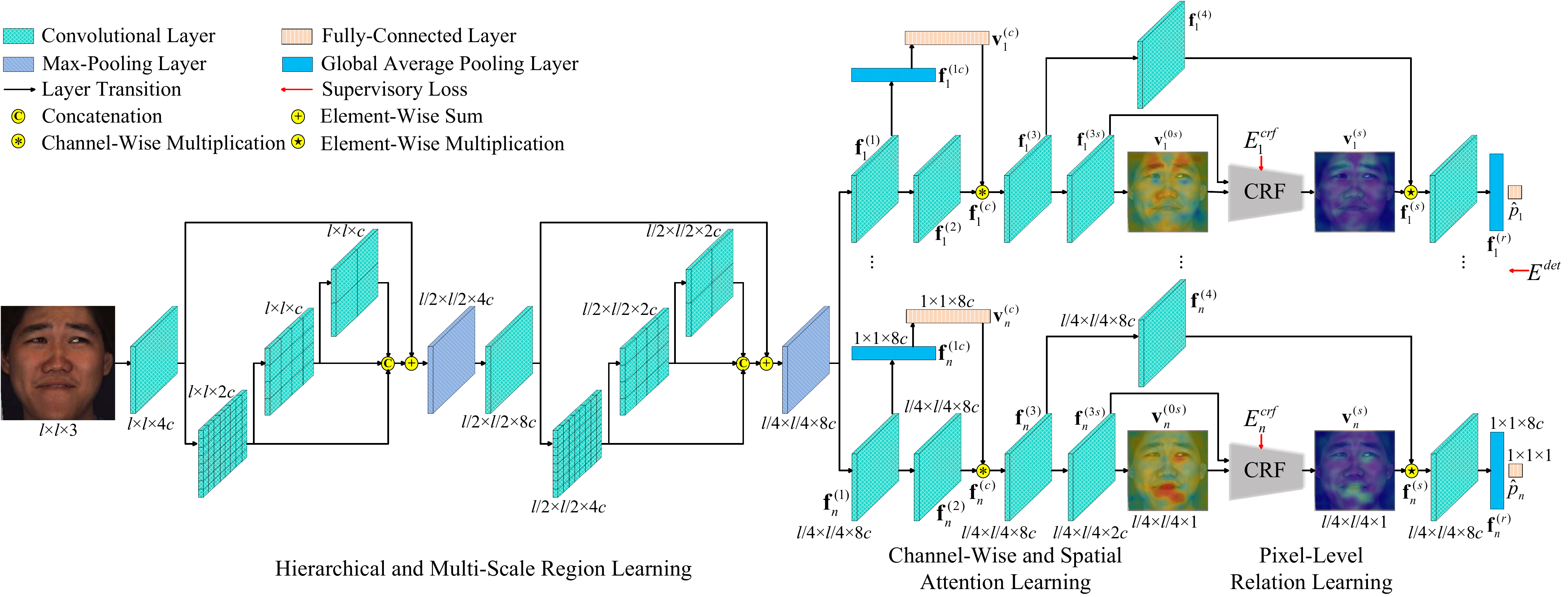}
\caption{Architecture of our ARL framework, in which channel-wise and spatial attention learning and pixel-level relation learning are the core modules for AU detection. The spatial attention weights $\mathbf{v}_1^{(s)}$ and $\mathbf{v}_n^{(s)}$ of two examples AU 4 and AU 17 are shown respectively, which are refined by the pixel-level relation learning with the attentions in irrelevant regions removed. The expression $l\times l\times c$ indicates that the height, width and channel of the corresponding layer are $l$, $l$ and $c$, respectively.}
\label{fig:attention_framework}
\end{figure*}

\subsection{Relation Based AU Detection}

Since AU detection is a multi-label classification problem, the relations among AUs can be exploited to improve the performance for each AU. Probabilistic Graphical Model (PGM)~\cite{pearl1988probabilistic} that combines probabilistic models and graphical models has been extensively used for the relation learning among AUs, due to its ability of modeling and visualization for complex data distribution. Tong et al.~\cite{tong2008learning} applied a Bayesian Network (BN) to model the local pairwise AU dependencies such as co-occurrence and mutual exclusion. Zhu et al.~\cite{zhu2014multiple} adopted a multi-task learning method for the recognition of multiple AU groups and employed the BN to calibrate the AU recognition results. Li et al.~\cite{li2013simultaneous} adopted a Dynamic Bayesian Network (DBN) to achieve facial landmark tracking, AU recognition and expression recognition simultaneously, in which the local AU relationships and local dependencies between AUs and landmarks are both captured. To overcome the pairwise AU modeling limitation of DBN, Wang et al.~\cite{wang2013capturing} introduced a Restricted Boltzmann Machine (RBM) to capture both local and global AU dependencies. BN, DBN and RBM are generative models, which cannot be jointly optimized with deep learning models in an end-to-end framework. In contrast, our method combines a deep convolutional neural network with the discriminative model CRF, which can be trained end-to-end using the back-propagation algorithm~\cite{rumelhart1985learning}.

By combining both generative and discriminative learning, Eleftheriadis et al.~\cite{eleftheriadis2015multi} proposed a multi-conditional latent variable model to jointly detect multiple AUs, in which both local and global relationships among AUs are encoded. Zhang et al.~\cite{zhang2016task} proposed a multi-task multiple kernel learning method with Support Vector Machine (SVM) classifiers to learn a kernel representation which encodes the AU relations. With a cascade regression framework, Wu et al.~\cite{wu2016constrained} captured global AU relationships and global dependencies between AUs and landmarks. Zhao et al.~\cite{zhao2016joint} proposed a Joint Patch and Multi-Label Learning (JPML) framework for AU detection by modeling the joint dependencies behind features, AUs, and their interplay. Recently, Corneanu et al.~\cite{corneanu2018deep} proposed a Deep Structure Inference Network (DSIN) to infer the structure among AUs through iterative message passing, which contains a collection of interconnected recurrent structure inference units. In contrast with these methods using AU-level relations, we propose to model the pixel-level relations for each AU, which are integrated with the attention mechanism.

\section{ARL for Facial AU Detection}
\subsection{Overview}

Our main goal is to predict the AU occurrence probabilities $\hat{\mathbf{p}}$ given an input face image with a size of $l\times l\times 3$, where $\hat{\mathbf{p}}=(\hat{p}_1, \cdots, \hat{p}_n)$ for all $n$ AUs. Fig.~\ref{fig:attention_framework} illustrates the architecture of our proposed framework ARL, which consists of three modules: hierarchical and multi-scale region learning, channel-wise and spatial attention learning, and pixel-level relation learning. Similar to~\cite{li2018eac,shao2018deep}, our framework uses stacked convolutional layers proposed by VGGNet \cite{simonyan2014very} to extract features. The module of hierarchical and multi-scale region learning aims to extract multi-scale features so as to adapt to various AUs with different sizes. Based on the shared multi-scale features, the modules of channel-wise and spatial attention learning and pixel-level relation learning adaptively capture attentions and relations for each AU, which are the central parts for AU detection. The main notations in our framework are summarized in Table~\ref{tab:notation}.

\begin{table}
\centering\caption{Notations in our framework.}
\label{tab:notation}
\begin{tabular}{|*{2}{c|}}
\hline
Notation &Definition\\\hline
$l$ &width of input image\\\hline
$c$ &structure hyperparameter of layer channels\\\hline
$\hat{p}_i$ &predicted occurrence probability of the $i$-th AU\\\hline
$p_i$ &ground-truth occurrence probability of the $i$-th AU\\\hline
$\hat{p}^{int}_i$ &predicted intensity of the $i$-th AU\\\hline
$p^{int}_i$ &ground-truth intensity of the $i$-th AU\\\hline
$\mathbf{f}_i^{(1)}$ &first feature of the $i$-th AU\\\hline
$\mathbf{f}_i^{(1c)}$ &channel-wise feature of the $i$-th AU\\\hline
$\mathbf{v}_i^{(c)}$ &channel-wise attention weight of the $i$-th AU\\\hline
$\mathbf{f}_i^{(2)}$ &second feature of the $i$-th AU\\\hline
$\mathbf{f}_i^{(c)}$ &channel-wise weighted feature of the $i$-th AU\\\hline
$\mathbf{f}_i^{(3)}$ &third feature of the $i$-th AU\\\hline
$\mathbf{f}_i^{(3s)}$ &spatial pixel-wise feature of the $i$-th AU\\\hline
$\mathbf{v}_i^{(0s)}$ &initial spatial attention weight of the $i$-th AU\\\hline
$\mathbf{v}_i^{(s)}$ &refined spatial attention weight of the $i$-th AU\\\hline
$\mathbf{f}_i^{(4)}$ &fourth feature of the $i$-th AU\\\hline
$\mathbf{f}_i^{(s)}$ &spatial pixel-wise weighted feature of the $i$-th AU\\\hline
$\mathbf{f}_i^{(r)}$ &finally related feature of the $i$-th AU\\\hline
$E^{crf}_i$ &CRF energy of the $i$-th AU\\\hline
$E^{cos}$ &cosine similarity loss\\\hline
$E^{det}$ &AU detection loss\\\hline
$E^{int}$ &AU intensity estimation loss\\\hline
\end{tabular}
\end{table}
\subsection{Hierarchical and Multi-Scale Region Learning}

We use the hierarchical and multi-scale region learning module in JAA-Net~\cite{shao2018deep}. It contains two blocks of the hierarchical and multi-scale region layer, each of which is followed by a max-pooling layer. Specifically, a hierarchical and multi-scale region layer includes an input layer and three hierarchical intermediate layers. The uniformly partitioned $8\times 8$, $4\times 4$ and $2\times 2$ patches of the three intermediate layers are the convolution outputs on corresponding patches in the input layer, the first intermediate layer and the second intermediate layer, respectively. Each local patch is processed with independent convolutional filters to extract local features. By concatenating the outputs of the three intermediate layers, we can extract hierarchical and multi-scale features, which are further summed element-wise with the output of the input layer. The element-wise sum operation is used as a residual structure~\cite{he2016deep} which is beneficial for avoiding the vanishing gradient problem. This module extracts multi-scale features to facilitate the further attention learning and relation learning for each AU.

\subsection{Channel-Wise and Spatial Attention Learning}

As shown in Fig.~\ref{fig:attention_framework}, each AU has the same structures of the attention learning and relation learning. For the $i$-th AU, $i=1,\cdots,n$, we apply a convolution operation on the output of the hierarchical and multi-scale region learning module to extract a feature $\mathbf{f}_i^{(1)}$. A channel-wise feature $\mathbf{f}_i^{(1c)}\in \mathbb{R}^{8c\times 1}$ and a feature $\mathbf{f}_i^{(2)}$ are generated by performing Global Average Pooling and convolution on $\mathbf{f}_i^{(1)}$, respectively. The \textit{channel-wise attention weight} $\mathbf{v}_i^{(c)}\in \mathbb{R}^{8c\times 1}$ is computed as
\begin{equation}
\mathbf{v}_i^{(c)} = \sigma({\mathbf{W}_i^{(1c)}}^T\mathbf{f}_i^{(1c)}),
\end{equation}
where $c$ is a hyperparameter with respect to the structure of our framework, $\mathbf{W}_i^{(1c)}\in \mathbb{R}^{8c\times 8c}$ denotes the weight parameters of the $8c$-dimensional fully-connected layer, and $\sigma(x)=1/(1+\exp(-x))$ is the sigmoid function. Without loss of generality, we simplify the notation of the linear mapping ${\mathbf{W}_i^{(1c)}}^T\mathbf{f}_i^{(1c)}$ by omitting the bias term. Then, we obtain the \textit{channel-wise weighted feature}:
\begin{equation}
\label{eq:channel}
\mathbf{f}_i^{(c)}= \mathbf{v}_i^{(c)}* \mathbf{f}_i^{(2)},
\end{equation}
where $*$ denotes the channel-wise multiplication of the feature map channels and the corresponding channel-wise attention weights. Considering each convolutional filter is analogous to performing a pattern detector~\cite{zhang2018interpretable} and a channel-wise feature map is the response of a certain filter, our proposed channel-wise attention learning is essentially selecting features with AU-related attributes.

During the process of spatial attention learning, a feature $\mathbf{f}_i^{(3)}$ is first generated by applying convolution on $\mathbf{f}_i^{(c)}$. Then we adopt a convolutional layer for $\mathbf{f}_i^{(3)}$ to extract a \textit{spatial pixel-wise feature} $\mathbf{f}_i^{(3s)}=(\mathbf{f}_{i1}^{(3s)}, \cdots, \mathbf{f}_{im}^{(3s)})$, which will be used for the learning of pixel-level relations. By processing $\mathbf{f}_i^{(3s)}$ with a one-channel convolutional layer and the sigmoid function $\sigma(\cdot)$, we obtain the \textit{initial spatial attention weight} $\mathbf{v}_i^{(0s)}=(v_{i1}^{(0s)}, \cdots, v_{im}^{(0s)})$. Note that $\mathbf{v}_i^{(0s)}$ corresponds to a downsampled input image with a size of $l/4\times l/4\times 3$, and $m=l/4\times l/4$ is the number of pixels.

\subsection{Pixel-Level Relation Learning}

In this section, we exploit CRF to model pixel-level relations so as to refine the initial spatial attention weight $\mathbf{v}_i^{(0s)}$. Learning spatial attentions is a pixel-wise binary classification problem. Denote $y_{ij}\in \{0,1\}$ as the attention label of the $j$-th pixel for the $i$-th AU, where $j=1,\cdots,m$. We use the fully-connected CRF model~\cite{krahenbuhl2011efficient,zheng2015conditional}, in which the energy of a label assignment $\mathbf{y}_i=(y_{i1}, \cdots, y_{im})$ is defined as
\begin{equation}
\label{eq:crf}
E^{crf}_i(\mathbf{y}_i)= \sum_j \psi_i^u(y_{ij}) + \sum_{j<k} \psi_i^p(y_{ij}, y_{ik}),
\end{equation}
where $\psi_i^u(y_{ij}) = -\log P^{(0)}(y_{ij})$ is the unary potential which measures the cost of assigning label $y_{ij}$ to the $j$-th pixel, $P^{(0)}(y_{ij})$ is the initial label assignment probability of the $j$-th pixel, and $\psi_i^p(y_{ij}, y_{ik})$ is the pairwise potential. In particular, $P^{(0)}(y_{ij}=1) = v_{ij}^{(0s)}$ and $P^{(0)}(y_{ij}=0) = 1-v_{ij}^{(0s)}$.

Previous fully-connected CRF works~\cite{krahenbuhl2011efficient,zheng2015conditional} use RGB values as a feature for each pixel to model $\psi_i^p(y_{ij}, y_{ik})$, which is difficult to capture AUs without distinct contour and texture. In contrast, we use the learned spatial pixel-wise feature $\mathbf{f}_i^{(3s)}$ for $\psi_i^p(y_{ij}, y_{ik})$, which is modeled with weighted Gaussian kernels:
\begin{equation}
\begin{split}
\label{eq:pair}
&\psi_i^p(y_{ij}, y_{ik}) = \mu_i (y_{ij}, y_{ik}) [w_i^{(1)} \exp(-\frac{\Vert \mathbf{o}_{j} - \mathbf{o}_{k}\Vert^2}{2\alpha_i^2}-\\
 &\frac{\Vert \mathbf{f}_{ij}^{(3s)} - \mathbf{f}_{ik}^{(3s)}\Vert^2}{2\beta_i^2})+ w_i^{(2)} \exp(-\frac{\Vert \mathbf{o}_{j} - \mathbf{o}_{k}\Vert^2}{2\gamma_i^2})],
\end{split}
\end{equation}
where $\mu_i(\cdot,\cdot)$ is the label compatibility function, $\mathbf{o}_{j}\in \mathbb{R}^{2\times 1}$ denotes the position vector with x- and y-coordinates of the $j$-th pixel, $\mathbf{f}_{ij}^{(3s)}\in \mathbb{R}^{2c\times 1}$, the hyperparameters $w_i^{(1)}$ and $w_i^{(2)}$ control the relative importance of two Gaussian kernels, and $\alpha_i$, $\beta_i$ and $\gamma_i$ control the scale of Gaussian kernels. The first kernel in Eq.~\ref{eq:pair} enforces nearby pixels with similar features to have the same label, and the second kernel is used to enforce smoothness. Eq.~\ref{eq:pair} can model the relationships between two pixels from both near regions and far regions.

We minimize the CRF energy $E^{crf}_i(\mathbf{y}_i)$ using iterative mean-field inference~\cite{krahenbuhl2011efficient}, which is further developed by~\cite{zheng2015conditional} through reformulating the iterative process as a Recurrent Neural Network (RNN). By using the RNN mean-field inference, we obtain the refined label assignment probability $P(y_{ij})$. The \textit{refined spatial attention weight} at each pixel is directly defined as
\begin{equation}
v_{ij}^{(s)} = P(y_{ij}=1).
\end{equation}
Thus, we obtain the refined spatial attention weight $\mathbf{v}_i^{(s)}=(v_{i1}^{(s)}, \cdots, v_{im}^{(s)})$. After that, the \textit{spatial pixel-wise weighted feature} is calculated as
\begin{equation}
\mathbf{f}_i^{(s)}= \mathbf{v}_i^{(s)}\star \mathbf{f}_i^{(4)},
\end{equation}
where $\star$ denotes the element-wise multiplication of each feature map channel and the spatial attention weight, and $\mathbf{f}_i^{(4)}$ is generated by applying convolution on $\mathbf{f}_i^{(3)}$. With the fully-connected CRF, our proposed pixel-level relation learning method can adaptively capture the pixel dependencies to refine spatial pixel-wise attentions. As shown in Fig.~\ref{fig:attention_framework}, the attentions in irrelevant regions are removed for the refined spatial attention weight, which contributes to extracting more accurate AU-related features.

\subsection{Overall Objective Function}

After the attention learning and relation learning, the finally learned AU-related feature $\mathbf{f}_i^{(r)}\in \mathbb{R}^{8c\times 1}$ is extracted by processing $\mathbf{f}_i^{(s)}$ with a convolutional layer and a Global Average Pooling layer. Note that the use of the convolutional layer and the Global Average Pooling layer further captures relations among regions by enlarging receptive fields. The estimated AU occurrence probability is computed as
\begin{equation}
\label{eq:probability}
\hat{p}_i = \sigma({\mathbf{w}_i^{(r)}}^T\mathbf{f}_i^{(r)}),
\end{equation}
where $\mathbf{w}_i^{(r)}\in \mathbb{R}^{8c\times 1}$ denotes the weight parameters of the last one-dimensional fully-connected layer. After sharing the multi-scale features outputted by the hierarchical and multi-scale region learning, each AU has an independent branch to predict its occurrence probability so as to supervise the learning of its attentions and relations.

Considering AU detection is a multi-label binary classification problem, we define the AU detection loss as a weighted cross entropy loss:
\begin{equation}
\label{eq:weight_loss}
E^{det}=-\sum_{i=1}^n w_i [p_i \log \hat{p}_i + (1-p_i) \log (1-\hat{p}_i)],
\end{equation}
where $p_i$ denotes the ground-truth occurrence probability of the $i$-th AU. To alleviate the data imbalance issue, we use the weight parameter $w_i$ to weight the loss of each AU following~\cite{shao2018deep}. We set $w_i = (1/r_i)/\sum_{k=1}^{n}(1/r_k)$, in which $r_i$ is the occurrence rate of the $i$-th AU in the training set. Incorporating the two losses $E^{det}$ and $E^{crf}_i$ from Eqs.~\ref{eq:weight_loss} and~\ref{eq:crf}, we yield the overall objective function of our ARL framework:
\begin{equation}
\min E_{ARL}= E^{det} + \sum_{i=1}^n E^{crf}_i.
\end{equation}
Our proposed framework is trainable end-to-end, in which the hierarchical and multi-scale region learning, channel-wise and spatial attention learning, and CRF based pixel-level relation learning are trained simultaneously.

Despite being devised for AU detection, our framework can be easily extended for AU intensity estimation, which requires the estimation of intensities rather than only occurrences of different AUs. In particular, the predicted intensity of the $i$-th AU is calculated as
\begin{equation}
\hat{p}^{int}_i = \hat{p}_i L,
\end{equation}
where $L$ denotes the maximum intensity level. The AU intensity estimation loss is defined as a weighted Euclidean loss:
\begin{equation}
\label{eq:weight_loss_int}
E^{int}=\frac{1}{2}\sum_{i=1}^n w_i (p^{int}_i-\hat{p}^{int}_i)^2,
\end{equation}
where $p^{int}_i$ denotes the ground-truth intensity of the $i$-th AU, and $1/2$ is for eliminating the coefficient of the derivative of $E^{int}$. We compute $w_i$ by treating the AU intensities of $\{\lceil L/2\rceil,\lceil L/2\rceil+1,\cdots,L\}$ as occurrences and $\{0,1,\cdots,\lceil L/2\rceil-1\}$ as non-occurrences in the training set.

Besides the distances, we also take the correlations between ground-truth and predicted AU intensities into account by using a cosine similarity loss:
\begin{equation}
\label{eq:cosine_loss}
E^{cos}=1 - \frac{\sum_{i=1}^n p^{int}_i\hat{p}^{int}_i}{\sqrt{\sum_{i=1}^n {p^{int}_i}^2}\sqrt{\sum_{i=1}^n \hat{p}{^{int}_i}^2} },
\end{equation}
where the second term is the cosine similarity between ground-truth and predicted AU intensities. Denote our ARL framework for intensity estimation as iARL. The overall objective function for AU intensity estimation is defined as
\begin{equation}
\label{eq:iARL}
\min E_{iARL}= E^{int} + \sum_{i=1}^n E^{crf}_i + \lambda E^{cos},
\end{equation}
where the hyperparameter $\lambda$ controls the importance of $E^{cos}$. Therefore, our framework can be naturally and easily extended for AU intensity estimation, without changing the network architecture.

During the inference process, we feed a given testing image into our ARL network to obtain the output $\hat{\mathbf{p}}=(\hat{p}_1, \cdots, \hat{p}_n)$. For the $i$-th AU, the predicted occurrence probability and intensity are discretized as $\lfloor \hat{p}_i\rceil\in\{0,1\}$ and $\lfloor \hat{p}^{int}_i\rceil\in \{0,1,\cdots,L\}$, respectively, in which $\lfloor \cdot \rceil$ denotes the operation of rounding a number to the nearest integer.

\begin{table*}
\centering\caption{F1-frame and accuracy results for $12$ AUs on BP4D. The best results are shown in bold and brackets, and the second best results are shown in bold. Note that only the F1-Frame results are reported by ATF~\cite{zhang2018identity} and DSIN~\cite{corneanu2018deep}. EAC-Net and JAA-Net are shortly written as EAC and JAA, respectively.}
\label{tab:comp_f1_acc_bp4d}
\begin{tabular}{|*{15}{c|}}
\hline
\multirow{2}*{AU} &\multicolumn{8}{c|}{F1-Frame} &\multicolumn{6}{c|}{Accuracy} \\
\cline{2-15}&LSVM&JPML&DRML&EAC&ATF&DSIN&JAA&\textbf{ARL}&LSVM&JPML&DRML&EAC&JAA&\textbf{ARL}\\\hline
1 &23.2 &32.6 &36.4 &39.0 &39.2 &[\textbf{51.7}] &\textbf{47.2} &45.8 &20.7 &40.7 &55.7 &68.9 &[\textbf{74.7}] &\textbf{73.9}\\
2 &22.8 &25.6 &\textbf{41.8} &35.2 &35.2 &40.4 &[\textbf{44.0}] &39.8 &17.7 &42.1 &54.5 &73.9 &[\textbf{80.8}] &\textbf{76.7}\\
4 &23.1 &37.4 &43.0 &48.6 &45.9 &[\textbf{56.0}] &54.9 &\textbf{55.1} &22.9 &46.2 &58.8 &78.1 &\textbf{80.4} &[\textbf{80.9}]\\
6 &27.2 &42.3 &55.0 &\textbf{76.1} &71.6 &\textbf{76.1} &[\textbf{77.5}] &75.7 &20.3 &40.0 &56.6 &\textbf{78.5} &[\textbf{78.9}] &78.2\\
7 &47.1 &50.5 &67.0 &72.9 &71.9 &73.5 &\textbf{74.6} &[\textbf{77.2}] &44.8 &50.0 &61.0 &69.0 &\textbf{71.0} &[\textbf{74.4}]\\
10 &77.2 &72.2 &66.3 &81.9 &79.0 &79.9 &[\textbf{84.0}] &\textbf{82.3} &73.4 &75.2 &53.6 &77.6 &[\textbf{80.2}] &\textbf{79.1}\\
12 &63.7 &74.1 &65.8 &86.2 &83.7 &85.4 &[\textbf{86.9}] &\textbf{86.6} &55.3 &60.5 &60.8 &84.6 &\textbf{85.4} &[\textbf{85.5}]\\
14 &64.3 &[\textbf{65.7}] &54.1 &58.8 &\textbf{65.5} &62.7 &61.9 &58.8 &46.8 &53.6 &57.0 &60.6 &[\textbf{64.8}] &\textbf{62.8}\\
15 &18.4 &38.1 &33.2 &37.5 &33.8 &37.3 &\textbf{43.6} &[\textbf{47.6}] &18.3 &50.1 &56.2 &78.1 &\textbf{83.1} &[\textbf{84.7}]\\
17 &33.0 &40.0 &48.0 &59.1 &60.0 &[\textbf{62.9}] &60.3 &\textbf{62.1} &36.4 &42.5 &50.0 &70.6 &\textbf{73.5} &[\textbf{74.1}]\\
23 &19.4 &30.4 &31.7 &35.9 &37.3 &38.8 &\textbf{42.7} &[\textbf{47.4}] &19.2 &51.9 &53.9 &81.0 &\textbf{82.3} &[\textbf{82.9}]\\
24 &20.7 &\textbf{42.3} &30.0 &35.8 &41.8 &41.6 &41.9 &[\textbf{55.4}] &11.7 &53.2 &53.9 &82.4 &\textbf{85.4} &[\textbf{85.7}]\\
\hline
Avg &35.3 &45.9 &48.3 &55.9 &55.4 &58.9 &\textbf{60.0} &[\textbf{61.1}] &32.2 &50.5 &56.0 &75.2 &[\textbf{78.4}] &\textbf{78.2}\\
\hline
\end{tabular}
\end{table*}

\section{Experiments}
\subsection{Datasets and Settings}

\subsubsection{Datasets}
\label{sssec:datasets}
Our framework is evaluated on four benchmarks: BP4D~\cite{zhang2014bp4d}, DISFA~\cite{mavadati2013disfa}, FERA 2015~\cite{valstar2015fera} and BP4D+~\cite{zhang2016multimodal}, in which BP4D, DISFA and BP4D+ are for AU detection, and FERA 2015, DISFA and BP4D+ are for AU intensity estimation. Each dataset is coded with FACS~\cite{ekman1997face} by certified experts. Note that we aim at frame-level prediction, and therefore other datasets such as CK+~\cite{lucey2010extended} are not used because they only have video-level annotations.
\begin{itemize}
\item \textbf{BP4D} contains $23$ female and $18$ male subjects associating with $8$ sessions, i.e. $328$ videos in total. These videos include about $140,000$ frames with AU occurrence labels, which are partitioned into subject-exclusive three folds~\cite{zhao2016deep}. Following the settings of previous works~\cite{zhao2016deep,li2018eac,corneanu2018deep,shao2018deep}, we conduct 3-fold cross-validation on $12$ AUs: 1, 2, 4, 6, 7, 10, 12, 14, 15, 17, 23 and 24, in which two folds are used for training and the remaining one is used for testing.
\item \textbf{DISFA} contains $27$ subjects with $12$ females and $15$ males, each of whom was recorded by a video with $4,845$ frames. Each frame was labeled with AU intensities on a six-point ordinal scale from $0$ to $5$. DISFA has more serious data imbalance problem than BP4D, in which most of the AUs have very low occurrence rates. For AU detection, according to the setting of DRML~\cite{zhao2016deep}, the AU intensities equal or greater than $2$ are considered as occurrences, otherwise considered as non-occurrences. Subject-exclusive 3-fold cross-validation is conducted on $8$ AUs: 1, 2, 4, 6, 9, 12, 25 and 26. For AU intensity estimation, following CCNN-IT~\cite{walecki2017deep}, we evaluate $12$ AUs (1, 2, 4, 5, 6, 9, 12, 15, 17, 20, 25 and 26) by training on $18$ subjects and testing on the remaining $9$ subjects.
\item \textbf{FERA 2015} employs the BP4D dataset to evaluate the AU intensity estimation, in which $5$ AUs (6, 10, 12, 14 and 17) are annotated with intensities ranging from $0$ to $5$. Following the setting of CCNN-IT~\cite{walecki2017deep}, the training partition with $21$ subjects and the development partition with $20$ subjects are used for training and testing, respectively.
\item \textbf{BP4D+} is a multimodal facial expression dataset with $140$ subjects ($58$ males and $82$ females) and $10$ sessions, including synchronized 2D, 3D, thermal and physiological data sequences. BP4D+ has larger scale and variability for images than BP4D. It provides AU annotations for each subject in $4$ sessions, with a total number of $197,875$ images. To conduct a cross-dataset test, we train our framework on the entire BP4D dataset with $41$ subjects, and test them on all the BP4D+ images. We report results of $12$ AUs and $5$ AUs for AU detection and AU intensity estimation, respectively.
\end{itemize}

\begin{table*}
\centering\caption{F1-frame and accuracy results for $8$ AUs on DISFA. The results of different AUs for most of the previous approaches fluctuate significantly due to the data imbalance issue.}
\label{tab:comp_f1_acc_disfa}
\begin{tabular}{|*{15}{c|}}
\hline
\multirow{2}*{AU} &\multicolumn{8}{c|}{F1-Frame} &\multicolumn{6}{c|}{Accuracy}\\
\cline{2-15}&LSVM&APL&DRML&EAC&ATF&DSIN&JAA&\textbf{ARL}&LSVM&APL&DRML&EAC&JAA&\textbf{ARL}\\
\hline
1 &10.8 &11.4 &17.3 &41.5 &[\textbf{45.2}] &42.4 &43.7 &\textbf{43.9} &21.6 &32.7 &53.3 &85.6 &[\textbf{93.4}] &\textbf{92.1}\\
2 &10.0 &12.0 &17.7 &26.4 &39.7 &39.0 &[\textbf{46.2}] &\textbf{42.1} &15.8 &27.8 &53.2 &84.9 &[\textbf{96.1}] &\textbf{92.7}\\
4 &21.8 &30.1 &37.4 &\textbf{66.4} &47.1 &[\textbf{68.4}] &56.0 &63.6 &17.2 &37.9 &60.0 &79.1 &\textbf{86.9} &[\textbf{88.5}]\\
6 &15.7 &12.4 &29.0 &[\textbf{50.7}] &\textbf{48.6} &28.6 &41.4 &41.8 &8.7 &13.6 &54.9 &69.1 &\textbf{91.4} &[\textbf{91.6}]\\
9 &11.5 &10.1 &10.7 &[\textbf{80.5}] &32.0 &\textbf{46.8} &44.7 &40.0 &15.0 &64.4 &51.5 &88.1 &\textbf{95.8} &[\textbf{95.9}]\\
12 &70.4 &65.9 &37.7 &[\textbf{89.3}] &55.0 &70.8 &69.6 &\textbf{76.2} &93.8 &[\textbf{94.2}] &54.6 &90.0 &91.2 &\textbf{93.9}\\
25 &12.0 &21.4 &38.5 &88.9 &86.4 &\textbf{90.4} &88.3 &[\textbf{95.2}] &3.4 &50.4 &45.6 &80.5 &\textbf{93.4} &[\textbf{97.3}]\\
26 &22.1 &26.9 &20.1 &15.6 &39.2 &42.2 &\textbf{58.4} &[\textbf{66.8}] &20.1 &47.1 &45.3 &64.8 &\textbf{93.2} &[\textbf{94.3}]\\
\hline
Avg &21.8 &23.8 &26.7 &48.5 &49.2 &53.6 &\textbf{56.0} &[\textbf{58.7}] &27.5 &46.0 &52.3 &80.6 &\textbf{92.7} &[\textbf{93.3}]\\
\hline
\end{tabular}
\end{table*}

\begin{table*}
\centering\caption{F1-frame and accuracy results for $12$ AUs of cross-dataset testing on BP4D+. The best results are shown in bold.}
\label{tab:bp4d_plus_AUocc}
\begin{tabular}{|*{15}{c|}}
\hline
\multicolumn{2}{|c|}{AU}&1&2&4&6&7&10&12&14&15&17&23&24&Avg\\
\hline
\multirow{2}*{F1-Frame}&JAA-Net&\textbf{34.0} &30.1 &22.3 &80.4 &76.3 &\textbf{88.0} &\textbf{87.9} &\textbf{65.8} &38.0 &43.5 &45.8 &29.1 &53.4\\
&\textbf{ARL}&29.9 &\textbf{33.1} &\textbf{27.1} &\textbf{81.5} &\textbf{83.0} &84.8 &86.2 &59.7 &\textbf{44.6} &\textbf{43.7} &\textbf{48.8} &\textbf{32.3} &\textbf{54.6}\\
\hline
\multirow{2}*{Accuracy}&JAA-Net&\textbf{77.4} &\textbf{84.6} &76.2 &79.9 &70.4 &\textbf{83.7} &\textbf{85.3} &\textbf{61.1} &86.5 &\textbf{76.5} &83.3 &87.3 &79.3\\
&\textbf{ARL}&67.2 &82.8 &\textbf{84.4} &\textbf{80.3} &\textbf{77.8} &80.7 &82.9 &59.1 &\textbf{88.0} &75.1 &\textbf{83.9} &\textbf{93.2} &\textbf{79.6}\\
\hline
\end{tabular}
\end{table*}

\begin{table*}
\centering\caption{AU occurrence rates (\%) in the training set of BP4D and DISFA datasets. ``-'' denotes the dataset does not contain this AU.}
\label{tab:au_occurrence}
\begin{tabular}{|*{16}{c|}}
\hline
AU &1 &2 &4 &6 &7 &9 &10 &12 &14 &15 &17 &23 &24 &25 &26\\
\hline
BP4D &21.1 &17.1 &20.3 &46.2 &54.9 &- &59.4 &56.2 &46.6 &16.9 &34.4 &16.5 &15.2 &- &-\\
\hline
DISFA &5.0 &4.0 &15.0 &8.1 &- &4.3 &- &13.2 &- &- &- &- &- &27.8 &8.9\\
\hline
\end{tabular}
\end{table*}

\subsubsection{Implementation Details}

Similar to JAA-Net~\cite{shao2018deep}, each face image is aligned to be $200\times 200\times 3$ using similarity transformation including rotation, uniform scaling, and translation, whose shape is preserved without facial expression changes. Before inputted to our ARL network, each face is further cropped into $176\times 176\times 3$ and horizontally flipped, randomly. In our network, all the convolutional layers use $3\times 3$ convolutional filters with a stride $1$ and a padding $1$, and all the max-pooling layers process $2\times 2$ spatial fields with a stride $2$. The structure hyperparameters are set with $l=176$ and $c=8$, and the maximum intensity level $L$ is $5$. Our network is trained for up to $12$ epochs using Caffe~\cite{jia2014caffe} with Stochastic Gradient Descent (SGD), a mini-batch size of $8$, a weight decay of 0.0005, and a Nesterov momentum~\cite{sutskever2013importance} of 0.9. The learning rate starts with $0.006$ and $0.0006$ for AU detection and AU intensity estimation respectively, and is multiplied by a factor of $0.3$ at every $2$ epochs.

For AU detection, following the settings of~\cite{li2018eac,corneanu2018deep,shao2018deep}, we train our network on DISFA with the parameters initialized from the well-trained BP4D model. Specifically, AUs 9, 25 and 26 of DISFA are not included in BP4D, which are initialized from correlated AUs 10, 23 and 24 based on the prior knowledge. Similar to~\cite{zheng2015conditional}, for the CRF based pixel-level relation learning, the parameters of the label compatibility function $\mu_i(\cdot,\cdot)$ are initialized using the Potts model~\cite{potts1952some}, and the number of iterations of the RNN mean-field inference is $5$. The hyperparameters $w_i^{(1)}$, $w_i^{(2)}$, $\alpha_i$, $\beta_i$ and $\gamma_i$ in Eq.~\ref{eq:pair} and $\lambda$ in Eq.~\ref{eq:iARL} are obtained by cross validation on a small set of training data: $w_i^{(1)}=5$, $w_i^{(2)}=3$, $\alpha_i=160$, $\beta_i=3$, $\gamma_i=3$ and $\lambda=0.5$.

\subsubsection{Evaluation Metrics}

The evaluation metrics for AU detection and AU intensity estimation are introduced below:
\begin{itemize}
\item \textbf{AU Detection:} We use two commonly used metrics, frame-based F1-score (F1-frame) and accuracy. F1-frame is defined as $F1=2PR/(P+R)$, where $P$ and $R$ denote precision and recall respectively. Note that for some AUs with very low occurrence rates, directly predicting them as non-occurrences can obtain high accuracy but extremely poor F1-frame results. These two metrics can measure the performance of AU detection comprehensively. The average results of F1-frame and accuracy over all AUs (Avg) are also reported, respectively. All the quantitative results are reported in percentage with \% omitted.
\item \textbf{AU Intensity Estimation:} We report two popular metrics, intra-class correlation (ICC(3,1))~\cite{shrout1979intraclass} and mean absolute error (MAE). ICC measures the reliability by considering both correlation and agreement between the predictions and AU intensity labels. MAE measures the difference by computing the average of absolute errors. Similarly, the average results of ICC and MAE over all AUs (Avg) are also shown.
\end{itemize}

\subsection{Comparison with State-of-the-Art Methods}

We compare our method against state-of-the-art methods under the same setting stated in Section~\ref{sssec:datasets}. For AU detection, we compare our ARL with LSVM~\cite{fan2008liblinear}, APL~\cite{zhong2015learning}, JPML~\cite{zhao2016joint}, DRML~\cite{zhao2016deep}, EAC-Net~\cite{li2018eac}, ATF~\cite{zhang2018identity}, DSIN~\cite{corneanu2018deep} and JAA-Net~\cite{shao2018deep}, in which JPML and DSIN are recent relation based works, and EAC-Net and JAA-Net are pioneering attention based works. Note that a few methods like CNN+LSTM~\cite{chu2017learning} and R-T1~\cite{li2017action} are not compared, since they process a sequence of images instead of a single image. For AU intensity estimation, we compare our iARL with MRF~\cite{sandbach2013markov}, LT-all~\cite{kaltwang2015latent}, OR-CNN~\cite{niu2016ordinal}, DRML~\cite{zhao2016deep},  COR-HIT~\cite{walecki2017copula},  CCNN-IT~\cite{walecki2017deep} and 2DC~\cite{linh2017deepcoder}. Several earlier methods are re-implemented by recent works using the same setting, in which the AU detection results of LSVM, APL, and JPML are reported in~\cite{zhao2016deep,li2018eac}, and the AU intensity estimation results of MRF, LT-all, OR-CNN and DRML are reported in~\cite{walecki2017copula,walecki2017deep}.

\begin{table*}
\centering\caption{ICC and MAE results for $5$ AUs on FERA 2015. Note that MRF, LT-all, COR-HIT and 2DC only report the ICC results.}
\label{tab:comp_intensity_fera}
\begin{tabular}{|*{13}{c|}}
\hline
\multirow{2}*{AU} &\multicolumn{8}{c|}{ICC (higher is better)} &\multicolumn{4}{c|}{MAE (lower is better)} \\
\cline{2-13}&MRF&LT-all&OR-CNN&DRML&COR-HIT&CCNN-IT&2DC&\textbf{iARL}&OR-CNN&DRML&CCNN-IT&\textbf{iARL}\\\hline
6 &0.72 &0.69 &0.60 &0.62 &[\textbf{0.76}] &\textbf{0.75} &[\textbf{0.76}] &0.72 &1.37 &1.37 &\textbf{1.14} &[\textbf{0.62}]\\
10 &\textbf{0.71} &0.58 &0.61 &0.64 &[\textbf{0.72}] &0.69 &\textbf{0.71} &[\textbf{0.72}] &1.39 &\textbf{1.25} &1.30 &[\textbf{0.69}]\\
12 &0.81 &0.76 &0.59 &0.74 &0.81 &[\textbf{0.86}] &\textbf{0.85} &\textbf{0.85} &1.37 &1.13 &\textbf{0.99} &[\textbf{0.51}]\\
14 &0.33 &0.30 &0.25 &0.31 &0.29 &0.40 &[\textbf{0.45}] &\textbf{0.44} &1.80 &\textbf{1.59} &1.65 &[\textbf{0.91}]\\
17 &0.30 &0.31 &0.31 &0.32 &0.36 &0.45 &\textbf{0.53} &[\textbf{0.57}] &1.19 &1.16 &\textbf{1.08} &[\textbf{0.55}]\\
\hline
Avg &0.58 &0.53 &0.47 &0.52 &0.59 &\textbf{0.63} &[\textbf{0.66}] &[\textbf{0.66}] &1.42 &1.30 &\textbf{1.23} &[\textbf{0.66}]\\
\hline
\end{tabular}
\end{table*}

\begin{table*}[!htb]
\centering\caption{ICC and MAE results for $12$ AUs on DISFA. Note that 2DC only reports the ICC results.}
\label{tab:comp_intensity_disfa}
\begin{tabular}{|*{10}{c|}}
\hline
\multirow{2}*{AU} &\multicolumn{5}{c|}{ICC (higher is better)} &\multicolumn{4}{c|}{MAE (lower is better)} \\
\cline{2-10}&OR-CNN&DRML&CCNN-IT&2DC&\textbf{iARL}&OR-CNN&DRML&CCNN-IT&\textbf{iARL}\\\hline
1 &0.03 &0.05 &\textbf{0.18} &[\textbf{0.70}] &0.13 &1.05 &\textbf{0.85} &0.87 &[\textbf{0.30}]\\
2 &0.07 &0.06 &0.15 &[\textbf{0.55}] &\textbf{0.36} &0.87 &0.70 &\textbf{0.63} &[\textbf{0.31}]\\
4 &0.01 &0.32 &0.61 &[\textbf{0.69}] &\textbf{0.68} &1.47 &1.07 &\textbf{0.86} &[\textbf{0.52}]\\
5 &0.00 &0.02 &\textbf{0.07} &0.05 &[\textbf{0.22}] &\textbf{0.17} &0.20 &0.26 &[\textbf{0.04}]\\
6 &0.29 &0.36 &[\textbf{0.65}] &\textbf{0.59} &0.56 &0.79 &0.75 &\textbf{0.73} &[\textbf{0.36}]\\
9 &0.08 &0.39 &\textbf{0.55} &[\textbf{0.57}] &0.36 &0.70 &0.58 &\textbf{0.57} &[\textbf{0.30}]\\
12 &0.67 &0.77 &0.82 &[\textbf{0.88}] &\textbf{0.86} &0.69 &0.59 &\textbf{0.55} &[\textbf{0.31}]\\
15 &0.13 &0.29 &\textbf{0.44} &0.32 &[\textbf{0.52}] &0.44 &0.47 &\textbf{0.38} &[\textbf{0.05}]\\
17 &\textbf{0.27} &0.19 &[\textbf{0.37}] &0.10 &[\textbf{0.37}] &0.59 &\textbf{0.57} &\textbf{0.57} &[\textbf{0.33}]\\
20 &0.00 &0.04 &[\textbf{0.28}] &0.08 &\textbf{0.12} &0.50 &0.48 &\textbf{0.45} &[\textbf{0.08}]\\
25 &0.59 &0.65 &0.77 &\textbf{0.90} &[\textbf{0.96}] &1.33 &1.36 &\textbf{0.81} &[\textbf{0.29}]\\
26 &0.33 &0.35 &[\textbf{0.54}] &0.50 &\textbf{0.60} &0.86 &0.77 &\textbf{0.64} &[\textbf{0.26}]\\
\hline
Avg &0.20 &0.29 &0.45 &[\textbf{0.50}] &\textbf{0.48} &0.79 &0.70 &\textbf{0.61} &[\textbf{0.26}]\\
\hline
\end{tabular}
\end{table*}

\subsubsection{AU Detection}

\noindent\textbf{Evaluation on BP4D.}
Table~\ref{tab:comp_f1_acc_bp4d} reports the F1-frame and accuracy results of our method ARL and state-of-the-art methods on BP4D benchmark. It can be observed that our proposed ARL outperforms all the state-of-the-art methods including relation based methods JPML and DSIN, and attention based methods EAC-Net, ATF and JAA-Net. Our ARL exploits pixel-level relations and obtains much better results than DSIN, which is the most recent work modeling AU-level relations. Compared to JAA-Net using both AU and landmark labels, our ARL achieves competitive performance with only AU labels, which demonstrates the effectiveness of our proposed attention and relation learning.

\noindent\textbf{Evaluation on DISFA.}
The F1-frame and accuracy results evaluated on DISFA are shown in Table~\ref{tab:comp_f1_acc_disfa}, where it can be seen that our ARL significantly outperforms all the previous works with large margins. In particular, ARL brings relative increments of $4.82\%$ and $0.65\%$ for average F1-frame and average accuracy over JAA-Net, respectively. Note that there is a severe data imbalance problem in DISFA than BP4D, which causes significant performance fluctuations for different AUs in most of the previous methods especially LSVM, APL and EAC-Net. In contrast, in addition to top results in average F1-frame and average accuracy, our ARL exhibits more stable performance for each AU.

\noindent\textbf{Evaluation on BP4D+.}
After performing within-dataset tests on BP4D and DISFA, we conduct a cross-dataset test on the large-scale BP4D+ when training on BP4D. Table~\ref{tab:bp4d_plus_AUocc} presents the F1-frame and accuracy results of JAA-Net and our ARL on BP4D+. Although there is a domain gap between BP4D and BP4D+, JAA-Net and our ARL both achieve good performance on BP4D+. Moreover, we can see that ARL outperforms JAA-Net in terms of both F1-frame and accuracy metrics, which demonstrates the better generalization ability of our method.

To conduct AU-level analysis, we show the AU occurrence rates in the training set averaged over 3 folds of BP4D and DISFA datasets in Table~\ref{tab:au_occurrence}, respectively. We can observe that the occurrence rates of AUs 6, 7, 10, 12 and 14 are much higher than those of other AUs for BP4D, and the occurrence rates of AUs 4, 12 and 25 are higher than others for DISFA. For the results of our ARL, we also can see that AUs 6, 7, 10 and 12 have higher F1-frame than other AUs in Table~\ref{tab:comp_f1_acc_bp4d}, and AUs 12 and 25 have the highest F1-frame in Table~\ref{tab:comp_f1_acc_disfa}. This is due to that the prediction results of other AUs with occurrence rates far below $50\%$ often strongly bias towards non-occurrence so as to cause low F1-frame. Note that there are exceptions for AU 14 which is worse than AU 17 in Table~\ref{tab:comp_f1_acc_bp4d}, and AU 4 which is worse than AU 26 in Table~\ref{tab:comp_f1_acc_disfa}. This demonstrates that the performance of different AUs is also influenced by other factors such as the use of weighting the loss of each AU in Eq.~\ref{eq:weight_loss}, and correlations among AUs.

\begin{table}
\centering\caption{ICC and MAE results for $5$ AUs of cross-dataset testing on BP4D+.}
\label{tab:bp4d_plus_intensity}
\begin{tabular}{|*{8}{c|}}
\hline
\multicolumn{2}{|c|}{AU}&6&10&12&14&17&Avg\\
\hline
\multirow{2}*{ICC}&JAA-Net&0.72 &\textbf{0.79} &\textbf{0.82} &0.14 &0.45 &0.59\\
&\textbf{iARL}&\textbf{0.78} &0.77 &\textbf{0.82} &\textbf{0.19} &\textbf{0.50} &\textbf{0.61}\\
\hline
\multirow{2}*{MAE}&JAA-Net&0.63 &0.59 &\textbf{0.63} &0.79 &0.38 &0.60\\
&\textbf{iARL}&\textbf{0.58} &\textbf{0.55} &0.70 &\textbf{0.71} &\textbf{0.33} &\textbf{0.57}\\
\hline
\end{tabular}
\end{table}

\begin{table}
\centering\caption{Structures of different variants of our framework for both AU detection and AU intensity estimation. \textbf{H}: Hierarchical and multi-scale region learning.
\textbf{W}: Weighting the loss of each AU. \textbf{C}: Channel-wise attention learning. \textbf{S}: Spatial attention learning. \textbf{P}: Pixel-level relation learning with $E^{crf}_i$.}
\label{tab:variant_f1}
\begin{tabular}{|*{10}{c|}}
\hline
Method&H&W&C&S&P&$E^{det}$&$E^{int}$&$E^{cos}$\\
\hline
B-Net &$\surd$ & & & & &$\surd$ & & \\
W-Net &$\surd$ &$\surd$ & & & &$\surd$ & & \\
WC-Net &$\surd$ &$\surd$ &$\surd$ & & &$\surd$ & & \\
WCS-Net &$\surd$ &$\surd$ &$\surd$ &$\surd$ & &$\surd$ & & \\
\textbf{ARL} &$\surd$ &$\surd$ &$\surd$ &$\surd$ &$\surd$ &$\surd$ & & \\
iB-Net &$\surd$ & & & & & &$\surd$ & \\
iW-Net &$\surd$ &$\surd$ & & & & &$\surd$ & \\
iWC-Net &$\surd$ &$\surd$ &$\surd$ & & & &$\surd$ & \\
iWCS-Net &$\surd$ &$\surd$ &$\surd$ &$\surd$ & & &$\surd$ & \\
iWCSP-Net &$\surd$ &$\surd$ &$\surd$ &$\surd$ &$\surd$ & &$\surd$ & \\
\textbf{iARL} &$\surd$ &$\surd$ &$\surd$ &$\surd$ &$\surd$ & &$\surd$ &$\surd$ \\
\hline
\end{tabular}
\end{table}

\subsubsection{AU Intensity Estimation}

\noindent\textbf{Evaluation on FERA 2015.}
The ICC and MAE results of different methods on FERA 2015 benchmark are shown in Table~\ref{tab:comp_intensity_fera}. It can be seen that our proposed iARL achieves higher ICC and lower MAE than state-of-the-art works. Note that AU intensity estimation is a multi-label ordinal classification problem, in which most of the state-of-the-art approaches including OR-CNN, COR-HIT, CCNN-IT and 2DC explicitly utilize ordinal models. Although designed for AU detection without using the ordinal model, our framework performs better than previous ordinal methods with an easy extension. Moreover, our iARL is superior to other deep learning based methods including OR-CNN, DRML, CCNN-IT and 2DC, which demonstrates the effectiveness of our framework.

\noindent\textbf{Evaluation on DISFA.} Table~\ref{tab:comp_intensity_disfa} shows the AU intensity estimation results for $12$ AUs on DISFA benchmark. It can be observed that our iARL significantly outperforms OR-CNN, DRML and CCNN-IT in terms of both ICC and MAE metrics. Without the use of the ordinal model, our iARL achieves comparable performance to the state-of-the-art ordinal method 2DC. The effective results on both FERA 2015 and DISFA indicate the generalization of our framework extended for AU intensity estimation.

\begin{table*}
\centering\caption{F1-frame and accuracy results for $12$ AUs of different variants of our ARL on BP4D. The best results are shown in bold.}
\label{tab:ablation_bp4d}
\begin{tabular}{|*{15}{c|}}
\hline
\multicolumn{2}{|c|}{AU}&1&2&4&6&7&10&12&14&15&17&23&24&Avg\\
\hline
\multirow{5}*{F1-Frame}&B-Net&44.1&35.6&48.7&71.6&72.5&77.4&83.8&55.8&42.7&59.1&41.0&50.5&56.9\\
&W-Net&\textbf{47.4}&31.4&49.5&71.6&73.4&79.4&85.0&57.4&45.0&60.2&43.9&48.5&57.7\\
&WC-Net&46.5&34.1&50.6&74.5&76.9&79.1&84.0&53.3&46.3&58.3&44.1&50.8&58.2\\
&WCS-Net&45.2&34.9&52.8&74.9&75.7&79.3&\textbf{86.6}&\textbf{58.9}&45.5&60.9&46.8&53.5&59.6\\
&\textbf{ARL}&45.8&\textbf{39.8}&\textbf{55.1}&\textbf{75.7}&\textbf{77.2}&\textbf{82.3}&\textbf{86.6}&58.8&\textbf{47.6}&\textbf{62.1}&\textbf{47.4}&\textbf{55.4}&\textbf{61.1}\\
\hline
\multirow{5}*{Accuracy}&B-Net&73.5&78.1&77.2&73.9&70.0&73.8&82.7&58.0&79.7&69.4&80.7&81.0&74.8\\
&W-Net&77.2&78.7&75.4&74.5&70.4&76.1&83.5&58.6&82.5&70.9&80.0&81.5&75.8\\
&WC-Net&\textbf{78.3}&\textbf{80.4}&76.4&76.0&73.7&75.7&82.8&58.9&83.8&69.5&82.3&82.6&76.7\\
&WCS-Net&73.4&78.9&\textbf{81.0}&76.7&73.7&77.2&85.0&61.8&81.6&71.1&\textbf{83.0}&83.8&77.3\\
&\textbf{ARL}&73.9&76.7&80.9&\textbf{78.2}&\textbf{74.4}&\textbf{79.1}&\textbf{85.5}&\textbf{62.8}&\textbf{84.7}&\textbf{74.1}&82.9&\textbf{85.7}&\textbf{78.2}\\
\hline
\end{tabular}
\end{table*}

\begin{table}
\centering\caption{ICC and MAE results for $5$ AUs of different variants of our iARL on FERA 2015.}
\label{tab:ablation_fera}
\begin{tabular}{|*{8}{c|}}
\hline
\multicolumn{2}{|c|}{AU}&6&10&12&14&17&Avg\\
\hline
\multirow{6}*{ICC}&iB-Net&0.66&0.67&0.80&0.42&0.50&0.61\\
&iW-Net&0.66&0.70&0.81&0.40&0.53&0.62\\
&iWC-Net&0.70&0.69&0.81&0.36&0.54&0.62\\
&iWCS-Net&0.71&0.68&0.82&0.39&0.52&0.63\\
&iWCSP-Net&\textbf{0.75}&\textbf{0.73}&0.84&0.36&\textbf{0.58}&0.65\\
&\textbf{iARL}&0.72&0.72&\textbf{0.85}&\textbf{0.44}&0.57&\textbf{0.66}\\
\hline
\multirow{6}*{MAE}&iB-Net&0.80&0.86&0.68&1.07&0.70&0.82\\
&iW-Net&0.80&0.83&0.67&1.05&0.65&0.80\\
&iWC-Net&0.71&0.75&0.57&1.21&0.63&0.77\\
&iWCS-Net&0.65&0.79&0.57&1.06&0.57&0.73\\
&iWCSP-Net&0.67&\textbf{0.68}&0.56&0.94&\textbf{0.52}&0.67\\
&\textbf{iARL}&\textbf{0.62}&0.69&\textbf{0.51}&\textbf{0.91}&0.55&\textbf{0.66}\\
\hline
\end{tabular}
\end{table}

\noindent\textbf{Evaluation on BP4D+.} Table~\ref{tab:bp4d_plus_intensity} shows the AU intensity estimation results of cross-dataset testing on BP4D+ when training on BP4D. We implement JAA-Net for AU intensity estimation by replacing its AU detection loss with our AU intensity estimation loss. We can observe that our iARL achieves higher average ICC and lower average MAE than JAA-Net. Compared to the results on FERA 2015 in Table~\ref{tab:comp_intensity_fera}, our iARL shows numerically comparable performance on BP4D+, especially for the MAE metric. This demonstrates that our approach has a good generalization ability.

\begin{figure}
\centering\includegraphics[width=0.98\linewidth]{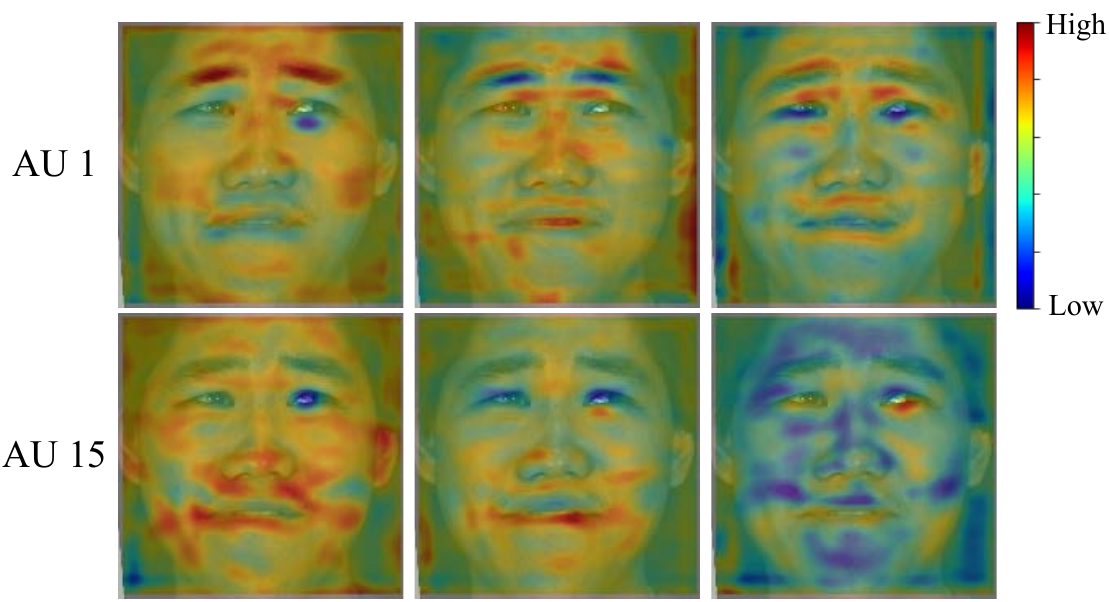}
\caption{Visualization of three feature map channels of $\mathbf{f}_i^{(2)}$ with the top three channel-wise attention weights in $\mathbf{v}_i^{(c)}$ for AUs 1 and 15 of the example sad face from BP4D, respectively. Colors from blue to red in the color bar indicate feature map values from low to high, which are overlaid on the face image.}
\label{fig:channel_attention}
\end{figure}

\subsection{Ablation Study}

In this section, we conduct experiments to evaluate the effectiveness of each component in our framework. Table~\ref{tab:variant_f1} summarizes the structures of different variants of our framework for both AU detection and AU intensity estimation. The baseline network B-Net consists of the hierarchical and multi-scale region learning, and five stacked convolutional layers, a Global Average Pooling layer and a one-dimensional fully-connected layer for each AU respectively. To investigate the effect of weighting the loss of each AU in Eq.~\ref{eq:weight_loss}, B-Net does not use the weighting method by setting each $w_i=1$. The structures of other variants are based on B-Net by replacing the corresponding parts with the proposed components. Note that B-Net and iB-Net are two baseline methods for AU detection and AU intensity estimation, respectively.

\begin{figure}
\centering\includegraphics[width=0.98\linewidth]{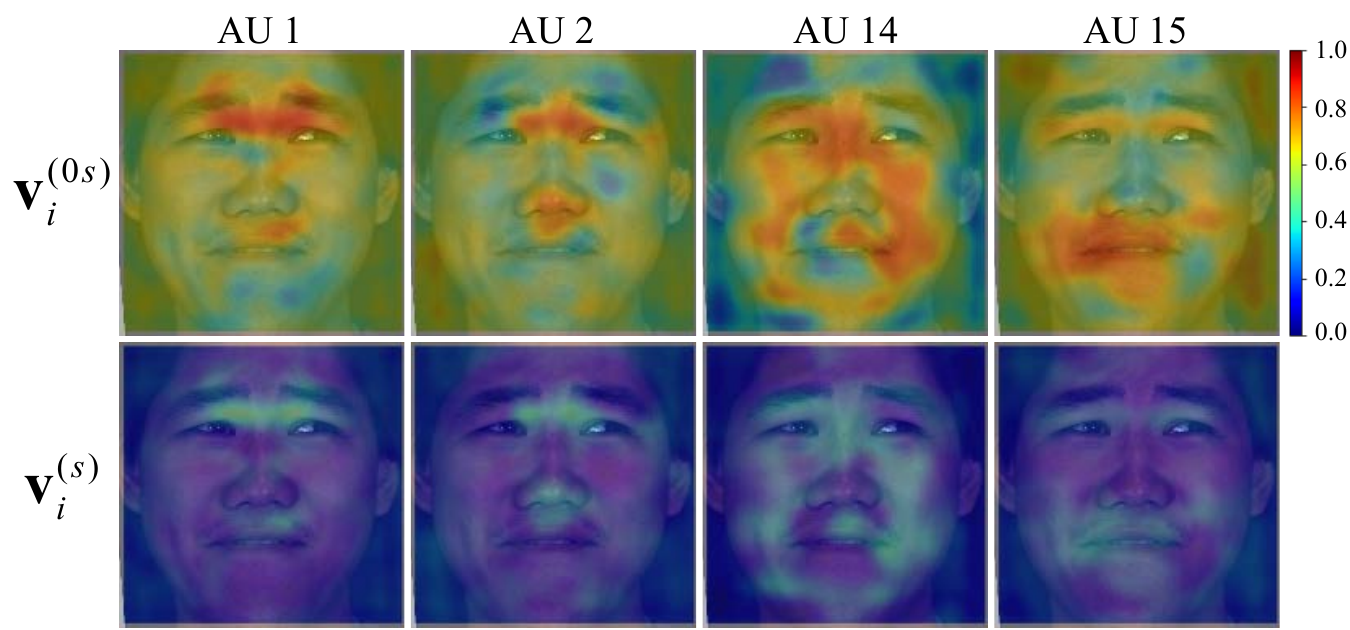}
\caption{Visualization of the initial spatial attention weight $\mathbf{v}_i^{(0s)}$ and refined spatial attention weight $\mathbf{v}_i^{(s)}$ for several AUs. Attention weights are visualized with the colors shown in the color bar.}
\label{fig:refine_attention}
\end{figure}

\begin{figure*}
\centering\includegraphics[width=0.98\linewidth]{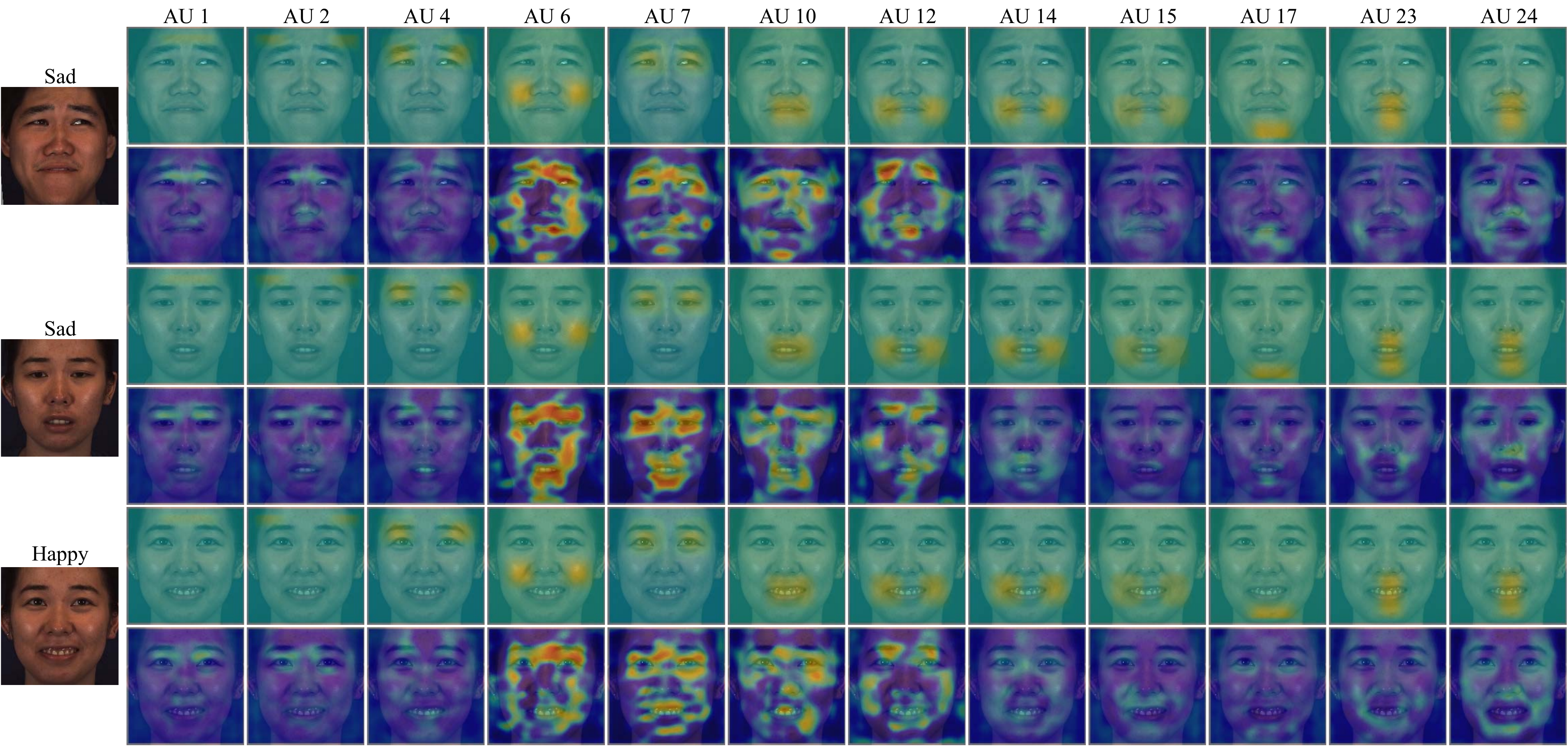}
\caption{Visualization of the refined spatial attention weights for 12 AUs of two sad faces and one happy face from BP4D, in which the second sad face and the third happy face belong to a different person compared with the first sad face. Every two rows show the results of the same image, where the first, third and fifth rows show the results of JAA-Net~\cite{shao2018deep}, and the second, fourth and sixth rows show the results of our method ARL.}
\label{fig:attention_map_bp4d}
\end{figure*}

\subsubsection{AU Detection}

Table~\ref{tab:ablation_bp4d} shows the F1-frame and accuracy results of different variants of our ARL on BP4D benchmark. It can be observed that W-Net performs better than B-Net, which demonstrates the effectiveness of weighting the loss of each AU. Based on W-Net, WC-Net advances the average F1-frame and average accuracy from levels 57 and 75 to the levels 58 and 76, respectively. WCS-Net further increases the average F1-frame and average accuracy to $59.6$ and $77.3$ by exploiting the spatial attention learning, respectively. It is verified that there are implicit attention mechanisms in deep neural networks (DNN)~\cite{zhao2016deep}. The channel-wise attention learning and spatial attention learning here explicitly learn channel-wise attentions and spatial attentions, and use them to weight DNN features.

After refining the spatial attention weight from $\mathbf{v}_i^{(0s)}$ to $\mathbf{v}_i^{(s)}$ using the pixel-level relation learning, our ARL achieves the best average F1-frame and average accuracy of $61.1$ and $78.2$, respectively. The large margins between the results of ARL and those of WC-Net are attributed to the integration of spatial attention learning and pixel-level relation learning. Note that the gaps of accuracy are smaller than those of F1-frame between different methods, such as $77.3-76.7=0.6$ of accuracy gap and $59.6-58.2=1.4$ of F1-frame gap between WCS-Net and WC-Net. This is because achieving a high F1-frame is more challenging than a high accuracy due to the data imbalance problem in BP4D dataset.

\subsubsection{AU Intensity Estimation}

The ICC and MAE results of different variants of our iARL on FERA 2015 are presented in Table~\ref{tab:ablation_fera}. We can see that the results become better after gradually integrating our proposed components into the structure of iB-Net, which indicates the effectiveness of our framework for AU intensity estimation. With weighting the loss of each AU, channel-wise attention learning, spatial attention learning and pixel-level relation learning, iWCSP-Net achieves good results with $0.65$ of average ICC and $0.67$ of average MAE. After exploiting the cosine similarity loss $E^{cos}$ to measure the correlations, our iARL further improves the ICC and MAE results. The uses of both $E^{int}$ and $E^{cos}$ contribute to training our iARL from the perspectives of both distances and correlations between ground-truth and predicted AU intensities.

\begin{figure*}
\centering\includegraphics[width=0.98\linewidth]{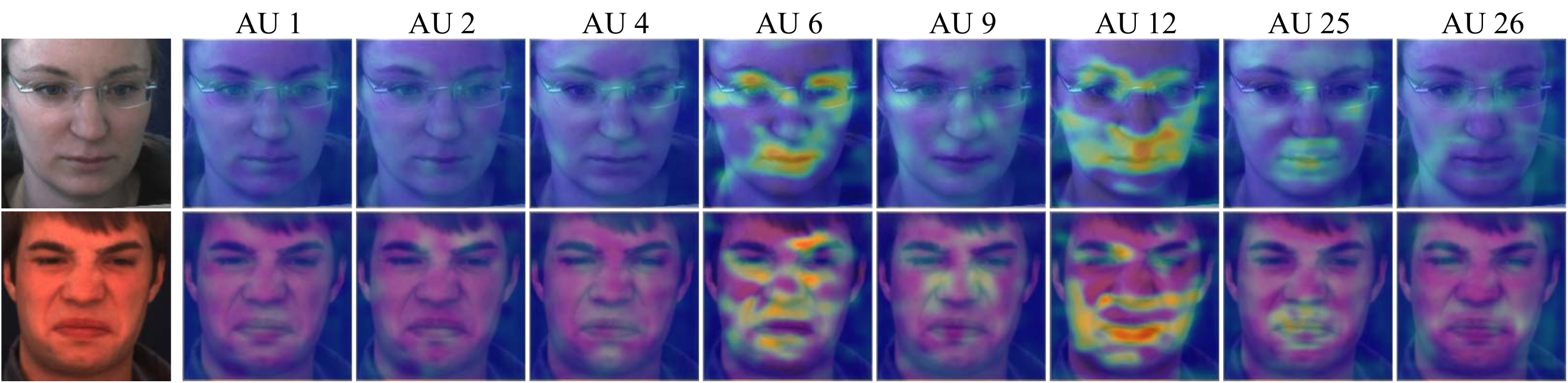}
\caption{Visualization of the refined spatial attention weight $\mathbf{v}_i^{(s)}$ of our ARL for $8$ AUs of two DISFA images. The first image is partially occluded by eyeglasses and the second image is in a different illumination environment.}
\label{fig:attention_map_disfa}
\end{figure*}

\subsection{Visual Results}

In this section, we show visual results with respect to features and attentions to validate our proposed framework.

\subsubsection{Channel-Wise Attention Learning}

Each feature map channel of the feature $\mathbf{f}_i^{(2)}$ is the response of a certain filter, which is weighted by the channel-wise attention weight $\mathbf{v}_i^{(c)}$. It is expected that $\mathbf{v}_i^{(c)}$ has larger weights for certain feature map channels of $\mathbf{f}_i^{(2)}$ generated by filters which capture AU-related information. To validate our proposed channel-wise attention learning, we visualize three feature map channels of $\mathbf{f}_i^{(2)}$ with the top three channel-wise attention weights in $\mathbf{v}_i^{(c)}$ for two examples AU 1 and AU 15 in Fig.~\ref{fig:channel_attention}, respectively. AU 1 denotes the inner brows raise and AU 15 denotes the lip corners depress, which usually co-occur in a sad expression~\cite{pantic2000expert}. It can be observed that the selected feature maps of AU 1 have highlights in the forehead and those of AU 15 have highlights in the mouth. This demonstrates that the channel-wise attention weight $\mathbf{v}_i^{(c)}$ can select features with related attributes for each AU from the feature $\mathbf{f}_i^{(2)}$. Note that the selected feature maps still contain other information such as the features of cheek and nose, and have noises in the background, which will be further processed by our learned spatial attention weights.

\subsubsection{Integration of Spatial Attention Learning and Pixel-Level Relation Learning}

To investigate the effect of our proposed pixel-level relation learning on spatial attentions, we visualize the initial spatial attention weight $\mathbf{v}_i^{(0s)}$ and refined spatial attention weight $\mathbf{v}_i^{(s)}$ for several AUs of the example sad face in Fig.~\ref{fig:refine_attention}. It can be seen that the spatial attention weights are refined with the attentions in irrelevant regions removed. Taking AU 2 (outer brows raise) as an example, the noisy attentions in the regions of facial contour and background are significantly reduced, and the correct attentions in the region of brow are preserved. Thus our proposed pixel-level relation learning is beneficial for capturing more accurate spatial attentions and extracting more relevant local features.

\subsubsection{Analysis of Spatial Attention Weight}

Here we illustrate that our learned spatial attention weight can adaptively capture the correlated regions of each AU. In particular, we visualize the refined spatial attention weights of ARL and JAA-Net~\cite{shao2018deep} for $12$ AUs of three example BP4D images in Fig.~\ref{fig:attention_map_bp4d}. There are two interesting observations from the visualization results as follows:
\begin{itemize}
\item The correlated regions of each AU should change across persons and facial expressions, which are difficult to be determined by the prior knowledge. This is because AUs are non-rigidly changed with facial expressions and the same expression of different persons is presented variously. The spatial attentions of AUs should be in irregular shape. It can be observed that the learned spatial attentions of JAA-Net are very similar for the same AU of different images, which are restricted by the prior knowledge. In contrast, our ARL adaptively learns spatial attentions under the supervision of AU detection and RNN mean-field inference. Specifically, each same AU has different spatial attention details for the first and second sad faces of two different persons, and each same AU also has different spatial attention details for the second sad face and the third happy face of a same person.
\item Besides the predefined ROIs by the prior knowledge, our learned spatial attentions are also highlighted in other correlated regions with adaptive responses. We can observe the implicit correlations among AUs from our learned spatial attentions. For instance, AUs 1 and 2, and AUs 6 and 7 have very similar attentions, which suggest the close relations between AUs 1 and 2, and between AUs 6 and 7. In addition, some AUs such as AU 6 which denotes the cheeks raise, are relevant to facial texture. They have slight or close correlations to more than one other AUs, whose spatial attentions are hard to be accurately learned by JAA-Net. Our learned spatial attentions have adaptive responses in all the correlated regions, in which the closely related regions have higher responses and the slightly related regions have lower responses.
\end{itemize}
Note that although a specific relationship between two AUs depends on a certain expression, we aim to model the relations among AUs in general cases. The correlations among AUs demonstrated by our learned spatial attentions are statistical laws in the training set. Fig.~\ref{fig:attention_map_disfa} visualizes the refined spatial attention weight $\mathbf{v}_i^{(s)}$ of our ARL for $8$ AUs of three DISFA images. It can be seen that our ARL can adaptively capture all the correlated regions for each AU. Moreover, ARL is robust to occlusion and illumination variations.

\begin{figure}
\centering\includegraphics[width=0.98\linewidth]{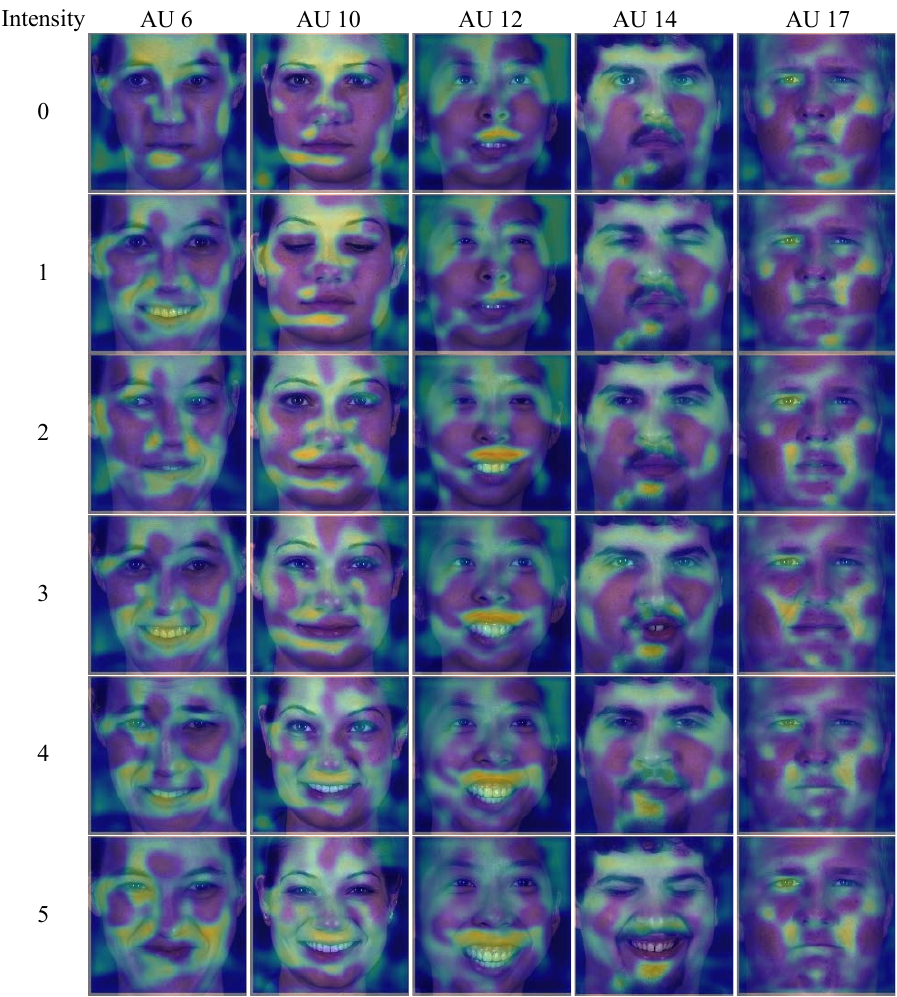}
\caption{Visualization of the $\mathbf{v}_i^{(s)}$ of our iARL for $5$ AUs on FERA 2015. Each column from top to bottom shows the results of FERA 2015 images of a same person with ground-truth AU intensities from $0$ to $5$.}
\label{fig:attention_intensity}
\end{figure}

\begin{figure*}
\centering\includegraphics[width=0.98\linewidth]{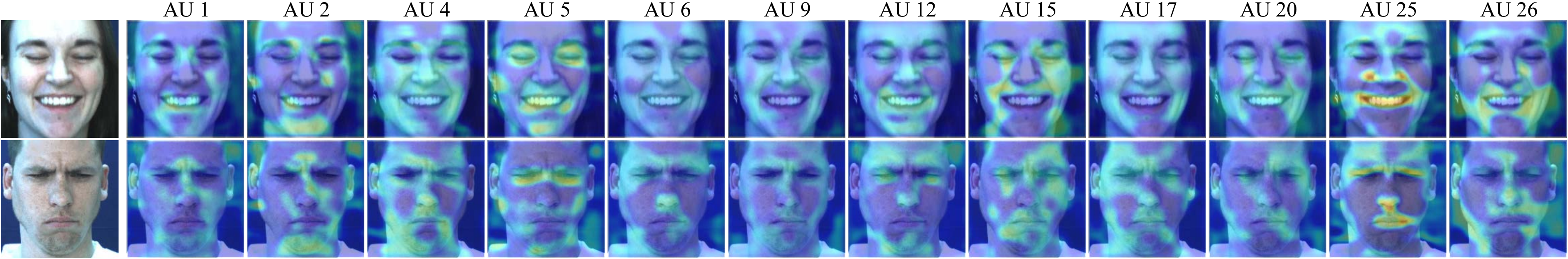}
\caption{Visualization of the $\mathbf{v}_i^{(s)}$ of our iARL for $12$ AUs of two example DISFA images.}
\label{fig:attention_map_intensity_disfa}
\end{figure*}

To validate our learned spatial attention weight for AU intensity estimation, we also visualize the $\mathbf{v}_i^{(s)}$ of our iARL for $5$ AUs on FERA 2015 in Fig.~\ref{fig:attention_intensity}. The showed AUs 6, 10, 12, 14, and 17 denote the cheeks raise, the upper lip raises, the lip corners pull, the dimples appear, and the chin raises respectively, whose ROIs are approximately the mouth and its neighboring regions. It can be seen that our learned spatial attentions for each AU have responses in both its ROI and other correlated regions, in which the responses in the mouth region of $0$ intensity are lower and narrower than those of intensities larger than $0$. This is because the estimation of $0$ intensity for each AU relies on more information from other correlated regions due to its non-occurrence in the mouth region. Moreover, comparing with the results of AU detection in Fig.~\ref{fig:attention_map_bp4d}, we find that the learned spatial attentions of AU intensity estimation occur in broader facial regions. This demonstrates that AU intensity estimation requires exploiting more information than AU detection in correlated regions to estimate the more detailed AU intensities. Fig.~\ref{fig:attention_map_intensity_disfa} visualizes the $\mathbf{v}_i^{(s)}$ of our iARL for two example DISFA images. We can observe that our iARL adaptively learns spatial attentions for each AU. With the proposed channel-wise and spatial attention learning and pixel-level relation learning, our framework can adapt to various AUs with different sizes and non-rigid transformations for both AU detection and AU intensity estimation.

\begin{figure}
\centering\includegraphics[width=0.98\linewidth]{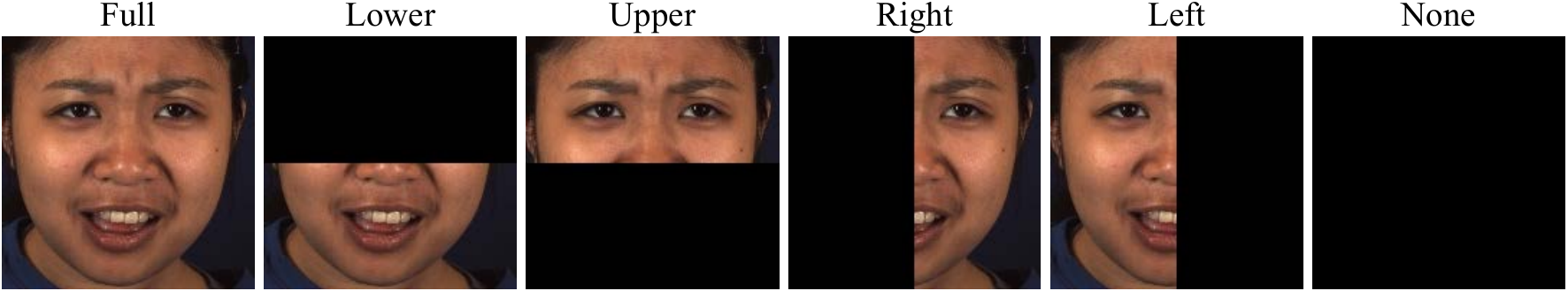}
\caption{An example BP4D face with different facial parts occluded. ``Full'' denotes the whole face is visible. ``None'' denotes the whole face is occluded.}
\label{fig:occlu_example}
\end{figure}

\begin{table}
\centering\caption{F1-frame results of EAC-Net and our ARL for $12$ AUs on partially occluded faces of BP4D. EAC-Net is shortly written as EAC.}
\label{tab:f1_occlu}
\begin{tabular}{|*{8}{c|}}
\hline
\multicolumn{2}{|c|}{AU} &Full &Lower &Upper &Right &Left &None\\\hline
\multirow{13}*{EAC}&1 &\textbf{39.0} &31.8 &27.4 &31.4 &25.2 &27.7\\
&2 &\textbf{35.2} &30.0 &31.6 &34.6 &32.4 &4.5\\
&4 &\textbf{48.6} &21.8 &29.1 &21.1 &28.7 &22.8\\
&6 &\textbf{76.1} &70.5 &54.9 &39.7 &52.9 &64.8\\
&7 &72.9 &72.3 &\textbf{74.4} &66.4 &70.1 &58.6\\
&10 &\textbf{81.9} &77.0 &64.6 &60.9 &62.6 &52.6\\
&12 &\textbf{86.2} &75.0 &67.6 &57.9 &59.9 &56.9\\
&14 &\textbf{58.8} &58.5 &51.6 &48.2 &45.7 &44.5\\
&15 &\textbf{37.5} &15.0 &14.8 &7.3 &18.4 &3.8\\
&17 &\textbf{59.1} &58.1 &39.3 &38.7 &37.8 &6.0\\
&23 &\textbf{35.9} &28.6 &18.9 &27.1 &12.5 &13.1\\
&24 &\textbf{35.8} &16.3 &13.0 &4.3 &7.6 &4.9\\
\cline{2-8}&Avg &\textbf{55.9} &46.3 &40.6 &36.5 &37.8 &29.8\\
\hline
\multirow{13}*{\textbf{ARL}}&1 &\textbf{45.8} &6.3 &20.4 &36.1 &43.3 &9.4\\
&2 &\textbf{39.8} &8.9 &17.1 &24.2 &38.5 &9.2\\
&4 &\textbf{55.1} &34.3 &17.1 &43.2 &44.7 &22.8\\
&6 &\textbf{75.7} &69.3 &57.1 &58.7 &46.8 &0.0\\
&7 &\textbf{77.2} &70.2 &65.6 &75.4 &59.9 &48.4\\
&10 &\textbf{82.3} &68.0 &56.3 &57.8 &66.4 &0.0\\
&12 &\textbf{86.6} &77.2 &68.6 &56.9 &55.3 &25.2\\
&14 &\textbf{58.8} &53.5 &36.3 &31.4 &35.7 &0.0\\
&15 &\textbf{47.6} &18.8 &18.0 &0.0 &0.1 &0.0\\
&17 &\textbf{62.1} &58.8 &51.5 &60.3 &58.3 &51.1\\
&23 &\textbf{47.4} &30.1 &12.1 &10.2 &26.3 &9.1\\
&24 &\textbf{55.4} &35.9 &10.4 &34.4 &45.7 &17.2\\
\cline{2-8}&Avg &\textbf{61.1} &44.3 &35.9 &40.7 &43.4 &16.0\\
\hline
\end{tabular}
\end{table}

\subsection{ARL for Partially Occluded Faces}

The correlations among AUs are very useful for AU detection especially for partially occluded faces. To investigate the influences of occlusions in different facial parts, we directly utilize our trained ARL model on BP4D without any additional processing to test partially occluded faces. Similar to~\cite{li2018eac}, these faces are occluded with only lower, upper, right and left half-faces visible, respectively. To confirm the results of partially occluded faces are better than randomly guessing, we also evaluate the test images with whole faces occluded, as shown in Fig.~\ref{fig:occlu_example}. Table~\ref{tab:f1_occlu} shows the F1-frame results of EAC-Net and our ARL on partially occluded faces of BP4D. Note that the prediction result of our ARL for the ``None'' face of each test image is fixed.

\begin{table}
\centering\caption{Overall F1-frame results on all the $9$ poses of FERA 2017. We compare our pARL with pEAC-Net and the baseline method of FERA 2017.}
\label{tab:f1_bp4d_9_pose}
\begin{tabular}{|*{4}{c|}}
\hline
AU&FERA 2017 baseline&pEAC-Net&\textbf{pARL}\\\hline
1 &15.4 &\textbf{27.2} &24.0\\
4 &17.2 &\textbf{33.2} &28.0\\
6 &56.4 &\textbf{69.9} &68.3\\
7 &72.7 &\textbf{80.8} &78.1\\
10 &69.2 &\textbf{83.4} &75.7\\
12 &64.7 &\textbf{80.2} &76.3\\
14 &62.2 &62.1 &\textbf{62.7}\\
15 &14.6 &25.1 &\textbf{30.0}\\
17 &22.4 &34.2 &\textbf{37.9}\\
23 &20.7 &26.1 &\textbf{39.8}\\
\hline
Avg &41.6 &\textbf{52.2} &52.1\\
\hline
\end{tabular}
\end{table}

We can see that our ARL achieves comparable performance to EAC-Net for partially occluded faces. Although EAC-Net performs better than ARL for images with only lower or upper half-faces visible, EAC-Net uses additional ground-truth landmark locations to predefine the ROI of each AU. The landmarks provide prior knowledge for prediction of AUs in the occluded half-face. On account of this, EAC-Net can obtain the average F1-frame of $29.8$ when whole faces are occluded. On the other hand, our ARL outperforms EAC-Net in terms of images with only right or left half-faces visible. This is due to that our ARL captures pixel-level relations including bilaterally symmetrical relations, in which the knowledge from the visible half-face can facilitate the prediction of AUs in the symmetrical occluded half-face.

\begin{figure}
\centering\includegraphics[width=0.98\linewidth]{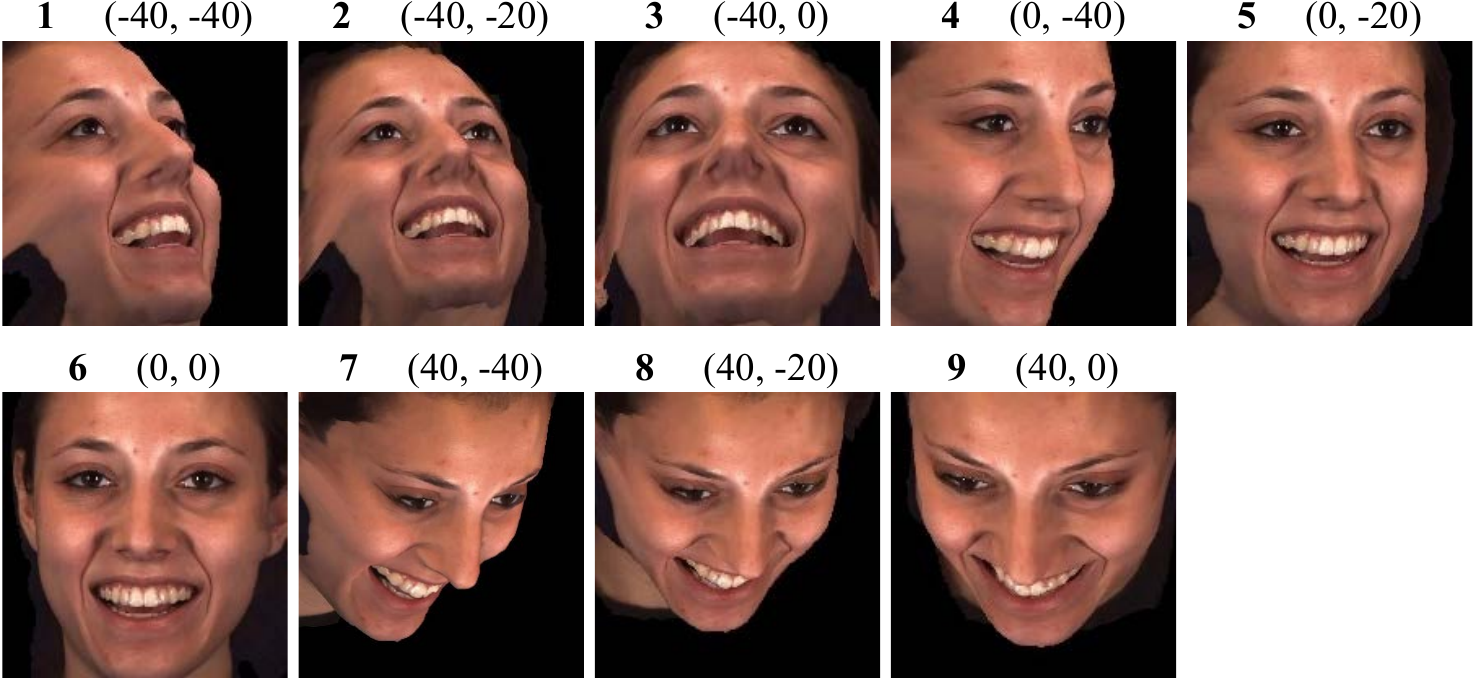}
\caption{Nine poses of an example face from FERA 2017. The angle degrees (yaw, pitch) of each pose are also shown.}
\label{fig:pose_example}
\end{figure}

\begin{table*}
\centering\caption{F1-frame results of pEAC-Net and our pARL for each of the $9$ poses of FERA 2017. pEAC-Net and pARL are shortly written as pE and pA, respectively.}
\label{tab:f1_bp4d_each_pose}
\begin{tabular}{|*{19}{c|}}
\hline
\multirow{2}*{AU} &\multicolumn{2}{c|}{1} &\multicolumn{2}{c|}{2} &\multicolumn{2}{c|}{3} &\multicolumn{2}{c|}{4} &\multicolumn{2}{c|}{5} &\multicolumn{2}{c|}{6} &\multicolumn{2}{c|}{7} &\multicolumn{2}{c|}{8} &\multicolumn{2}{c|}{9} \\
\cline{2-19}&pE&\textbf{pA}&pE&\textbf{pA}&pE&\textbf{pA}&pE&\textbf{pA}&pE&\textbf{pA}&pE&\textbf{pA}&pE&\textbf{pA}&pE&\textbf{pA}&pE&\textbf{pA}\\\hline
1 &18.8 &\textbf{22.9}&\textbf{37.5} &26.3 &22.2 &\textbf{24.2} &10.1 &\textbf{25.3} &\textbf{40.0} &26.3 &\textbf{66.7} &26.3 &\textbf{40.1} &16.9 &\textbf{33.3} &21.0 &\textbf{40.0} &24.0\\
4 &7.0 &\textbf{23.1} &\textbf{40.0} &24.6 &\textbf{33.3} &29.2 &\textbf{40.0} &29.6 &\textbf{30.7} &30.0 &\textbf{66.7} &30.7 &\textbf{25.0} &22.7 &18.2 &\textbf{31.3} &8.0 &\textbf{30.8}\\
6 &\textbf{66.7} &59.8 &66.7 &\textbf{68.0} &\textbf{72.2} &71.5 &60.8 &\textbf{70.6} &68.2 &\textbf{70.0} &\textbf{80.0} &71.5 &\textbf{69.2} &57.9 &\textbf{75.0} &69.8 &52.1 &\textbf{70.5}\\
7 &66.7 &\textbf{70.7} &62.9 &\textbf{79.5} &73.1 &\textbf{80.8} &75.7 &\textbf{79.6} &\textbf{85.7} &80.5 &\textbf{85.7} &81.0 &\textbf{83.7} &72.7 &\textbf{84.7} &76.2 &73.3 &\textbf{78.2}\\
10 &\textbf{85.0} &65.4 &69.2 &\textbf{78.4} &76.9 &\textbf{78.2} &\textbf{83.3} &78.8 &\textbf{90.9} &79.1 &75.0 &\textbf{79.2} &62.8 &\textbf{66.4} &\textbf{92.3} &75.3 &\textbf{82.8} &75.4\\
12 &\textbf{74.3} &67.2 &68.5 &\textbf{78.1} &\textbf{85.1} &78.6 &74.2 &\textbf{79.5} &\textbf{86.7} &80.3 &66.7 &\textbf{81.3} &64.3 &\textbf{64.8} &\textbf{84.2} &76.1 &72.0 &\textbf{76.2}\\
14 &\textbf{66.7} &58.5 &51.1 &\textbf{63.6} &51.2 &\textbf{60.9} &\textbf{68.9} &62.0 &59.3 &\textbf{64.4} &50.0 &\textbf{62.8} &\textbf{60.0} &58.0 &61.2 &\textbf{65.0} &\textbf{74.1} &67.2\\
15 &\textbf{57.1} &13.9 &5.1 &\textbf{31.7} &\textbf{36.4} &\textbf{36.4} &25.0 &\textbf{31.9} &28.6 &\textbf{34.1} &10.2 &\textbf{35.1} &7.0 &\textbf{12.3} &\textbf{40.0} &30.3 &11.8 &\textbf{32.8}\\
17 &23.5 &\textbf{34.7} &36.4 &\textbf{40.4} &34.7 &\textbf{40.9} &35.3 &\textbf{39.8} &\textbf{46.2} &39.9 &\textbf{50.0} &40.7 &\textbf{58.8} &29.5 &21.4 &\textbf{34.6} &\textbf{44.4} &36.2\\
23 &\textbf{44.4} &28.0 &36.3 &\textbf{40.8} &7.3 &\textbf{40.3} &\textbf{57.1} &44.8 &\textbf{47.1} &46.2 &6.3 &\textbf{43.7} &9.0 &\textbf{31.0} &\textbf{46.2} &39.1 &11.7 &\textbf{37.4}\\
\hline
Avg &\textbf{50.3} &44.4 &46.9 &\textbf{53.1} &48.5 &\textbf{54.1} &52.1 &\textbf{54.2} &\textbf{58.3} &55.1 &54.1 &\textbf{55.2} &\textbf{46.4} &43.2 &\textbf{55.7} &51.9 &46.2 &\textbf{52.9}\\
\hline
\end{tabular}
\end{table*}

We also notice that the result of each AU using our ARL becomes worse even if its predefined ROI is not occluded. This is because the prediction of a certain AU requires information from correlated AUs in other facial parts which might be occluded. Besides, the lower half-face performs best for AU detection, according to the results of both EAC-Net and ARL.  The reason is that the predefined ROIs of most AUs including AUs 10, 12, 14, 15, 17, 23 and 24 are in the lower half of a face. In some cases that upper half-faces including eyes are occluded, we can exploit knowledge from the lower facial part especially the mouth region for AU detection.

\subsection{ARL for Faces with Wide Pose Variations}

In this section, we evaluate our framework on faces with a wide range of head poses. FERA 2017~\cite{valstar2017fera} consists of BP4D~\cite{zhang2014bp4d} and BP4D+~\cite{zhang2016multimodal} with $9$ poses, in which the 3D model of each face is rotated by pitch angles of $-40$, $-20$ and $0$ degrees, and yaw angles of $-40$, $0$ and $40$ degrees from its frontal pose, respectively. Each face is annotated with 10 AUs: 1, 4, 6, 7, 10, 12, 14, 15, 17 and 23. Fig.~\ref{fig:pose_example} illustrates the nine poses of an example face. FERA 2017 utilizes BP4D with $41$ subjects for training, a subset of BP4D+ with $20$ subjects for validation, and a subset of BP4D+ with $30$ subjects for testing. We only report results on the validation set following pEAC-Net~\cite{li2018eac}. Denote our ARL framework for faces with wide pose variations as pARL. We compare our pARL against state-of-the-art pEAC-Net and the baseline method of FERA 2017.

We show the overall F1-frame results on all the $9$ poses in Table~\ref{tab:f1_bp4d_9_pose}. It can be observed that our pARL achieves competitive performance compared to the state-of-the-art pEAC-Net. Note that since pose is highly related to facial landmarks, pEAC-Net reduces the challenge of large poses by using ground-truth landmark locations. Without using additional landmark information, our pARL performs well for faces with a wide range of poses. We also list the F1-frame results of pEAC-Net and our pARL for each pose of FERA 2017 respectively in Table~\ref{tab:f1_bp4d_each_pose}. It can be seen that our pARL is robust to pose changes, in which the average F1-frame results of other poses are close to $55.2$ of pose 6 that is frontal with zero degrees of yaw and pitch angles. For example, our pARL achieves the average F1-frame of $54.2$ and $55.1$ for pose 4 and pose 5, respectively. In contrast, the average F1-frame results of pEAC-Net for pose 4 and pose 5 are far from $54.1$ of pose 6. Without the aid of landmark information and prior knowledge, our pARL still achieves good performance for faces with large poses such as poses 2, 3, 4 and 9.

\section{Conclusion}

In this paper, we have proposed an end-to-end deep learning based attention and relation learning framework for AU detection with only AU labels. Both channel-wise attention learning and spatial attention learning are used to select and extract AU-related local features. Moreover, we have proposed the pixel-level relation learning to refine spatial attentions so as to extract more accurate AU-related features. We have further extended our framework for AU intensity estimation without changing the network architecture.

We have compared our proposed framework against state-of-the-art methods on the challenging BP4D, DISFA, FERA 2015 and BP4D+ benchmarks. The experimental results demonstrate that our framework soundly outperforms the state-of-the-art methods for both AU detection and AU intensity estimation, and works well when cross-dataset testing on large-scale images. In addition, each component of our framework is indicated to be beneficial for both AU detection and AU intensity estimation. Extensive visual results demonstrate that our proposed channel-wise and spatial attention learning and pixel-level relation learning are beneficial for extracting AU-related features, and can adaptively capture the correlated regions of each AU.

We have further evaluated our framework on partially occluded faces of BP4D and demonstrated that our trained ARL model can be directly used to detect AUs with only half-faces visible. In addition, we have validated that our framework performs well on faces with a wide range of poses including large poses. Without exploiting landmark information, our method achieves comparable performance to EAC-Net which utilizes ground-truth landmark locations to predefine the ROI of each AU. We believe that the idea of joint learning of attention and pixel-level relation is also promising for other face analysis tasks such as facial expression recognition and face recognition.


%

%

\ifCLASSOPTIONcompsoc
  \section*{Acknowledgments}
\else
  \section*{Acknowledgment}
\fi

This work was supported by the National Natural Science Foundation of China (No. 61503277 and No. 61472245), the National Social Science Foundation of China (No. 18ZD22), and the Science and Technology Commission of Shanghai Municipality Program (No. 18D1205903). It was also partially supported by Data Science \& Artificial Intelligence Research Centre@NTU (DSAIR) and SINGTEL-NTU Cognitive \& Artificial Intelligence Joint Lab (SCALE@NTU), and the joint project of Tencent YouTu and Shanghai Jiao Tong University.

\ifCLASSOPTIONcaptionsoff
  \newpage
\fi



\bibliographystyle{IEEEtran}
\bibliography{references}

\begin{thebibliography}{10}
\providecommand{\url}[1]{#1}
\csname url@samestyle\endcsname
\providecommand{\newblock}{\relax}
\providecommand{\bibinfo}[2]{#2}
\providecommand{\BIBentrySTDinterwordspacing}{\spaceskip=0pt\relax}
\providecommand{\BIBentryALTinterwordstretchfactor}{4}
\providecommand{\BIBentryALTinterwordspacing}{\spaceskip=\fontdimen2\font plus
\BIBentryALTinterwordstretchfactor\fontdimen3\font minus
  \fontdimen4\font\relax}
\providecommand{\BIBforeignlanguage}[2]{{%
\expandafter\ifx\csname l@#1\endcsname\relax
\typeout{** WARNING: IEEEtran.bst: No hyphenation pattern has been}%
\typeout{** loaded for the language `#1'. Using the pattern for}%
\typeout{** the default language instead.}%
\else
\language=\csname l@#1\endcsname
\fi
#2}}
\providecommand{\BIBdecl}{\relax}
\BIBdecl

\bibitem{ekman1997face}
P.~Ekman and E.~L. Rosenberg, \emph{What the face reveals: Basic and applied
  studies of spontaneous expression using the Facial Action Coding System
  (FACS)}.\hskip 1em plus 0.5em minus 0.4em\relax Oxford University Press, USA,
  1997.

\bibitem{kuen2016recurrent}
J.~Kuen, Z.~Wang, and G.~Wang, ``Recurrent attentional networks for saliency
  detection,'' in \emph{IEEE Conference on Computer Vision and Pattern
  Recognition}.\hskip 1em plus 0.5em minus 0.4em\relax IEEE, 2016, pp.
  3668--3677.

\bibitem{liu2018picanet}
N.~Liu, J.~Han, and M.-H. Yang, ``Picanet: Learning pixel-wise contextual
  attention for saliency detection,'' in \emph{IEEE Conference on Computer
  Vision and Pattern Recognition}.\hskip 1em plus 0.5em minus 0.4em\relax IEEE,
  2018, pp. 3089--3098.

\bibitem{xiao2015application}
T.~Xiao, Y.~Xu, K.~Yang, J.~Zhang, Y.~Peng, and Z.~Zhang, ``The application of
  two-level attention models in deep convolutional neural network for
  fine-grained image classification,'' in \emph{IEEE Conference on Computer
  Vision and Pattern Recognition}.\hskip 1em plus 0.5em minus 0.4em\relax IEEE,
  2015, pp. 842--850.

\bibitem{cao2015look}
C.~Cao, X.~Liu, Y.~Yang, Y.~Yu, J.~Wang, Z.~Wang, Y.~Huang, L.~Wang, C.~Huang,
  W.~Xu, D.~Ramanan, and T.~S. Huang, ``Look and think twice: Capturing
  top-down visual attention with feedback convolutional neural networks,'' in
  \emph{IEEE International Conference on Computer Vision}.\hskip 1em plus 0.5em
  minus 0.4em\relax IEEE, 2015, pp. 2956--2964.

\bibitem{you2016image}
Q.~You, H.~Jin, Z.~Wang, C.~Fang, and J.~Luo, ``Image captioning with semantic
  attention,'' in \emph{IEEE Conference on Computer Vision and Pattern
  Recognition}, 2016, pp. 4651--4659.

\bibitem{pedersoli2017areas}
M.~Pedersoli, T.~Lucas, C.~Schmid, and J.~J. Verbeek, ``Areas of attention for
  image captioning,'' in \emph{IEEE International Conference on Computer
  Vision}.\hskip 1em plus 0.5em minus 0.4em\relax IEEE, 2017, pp. 1242--1250.

\bibitem{shao2018deep}
Z.~Shao, Z.~Liu, J.~Cai, and L.~Ma, ``Deep adaptive attention for joint facial
  action unit detection and face alignment,'' in \emph{European Conference on
  Computer Vision}.\hskip 1em plus 0.5em minus 0.4em\relax Springer, 2018, pp.
  725--740.

\bibitem{li2018eac}
W.~Li, F.~Abtahi, Z.~Zhu, and L.~Yin, ``Eac-net: Deep nets with enhancing and
  cropping for facial action unit detection,'' \emph{IEEE Transactions on
  Pattern Analysis and Machine Intelligence}, vol.~40, no.~11, pp. 2583--2596,
  2018.

\bibitem{li2017action}
W.~Li, F.~Abtahi, and Z.~Zhu, ``Action unit detection with region adaptation,
  multi-labeling learning and optimal temporal fusing,'' in \emph{IEEE
  Conference on Computer Vision and Pattern Recognition}.\hskip 1em plus 0.5em
  minus 0.4em\relax IEEE, 2017, pp. 6766--6775.

\bibitem{pantic2000expert}
M.~Pantic and L.~J. Rothkrantz, ``Expert system for automatic analysis of
  facial expressions,'' \emph{Image and Vision Computing}, vol.~18, no.~11, pp.
  881--905, 2000.

\bibitem{tong2008learning}
Y.~Tong and Q.~Ji, ``Learning bayesian networks with qualitative constraints,''
  in \emph{IEEE Conference on Computer Vision and Pattern Recognition}.\hskip
  1em plus 0.5em minus 0.4em\relax IEEE, 2008, pp. 1--8.

\bibitem{li2013simultaneous}
Y.~Li, S.~Wang, Y.~Zhao, and Q.~Ji, ``Simultaneous facial feature tracking and
  facial expression recognition,'' \emph{IEEE Transactions on Image
  Processing}, vol.~22, no.~7, pp. 2559--2573, 2013.

\bibitem{wu2016constrained}
Y.~Wu and Q.~Ji, ``Constrained joint cascade regression framework for
  simultaneous facial action unit recognition and facial landmark detection,''
  in \emph{IEEE Conference on Computer Vision and Pattern Recognition}.\hskip
  1em plus 0.5em minus 0.4em\relax IEEE, 2016, pp. 3400--3408.

\bibitem{zhang2016task}
X.~Zhang and M.~H. Mahoor, ``Task-dependent multi-task multiple kernel learning
  for facial action unit detection,'' \emph{Pattern Recognition}, vol.~51, pp.
  187--196, 2016.

\bibitem{wang2013capturing}
Z.~Wang, Y.~Li, S.~Wang, and Q.~Ji, ``Capturing global semantic relationships
  for facial action unit recognition,'' in \emph{IEEE International Conference
  on Computer Vision}.\hskip 1em plus 0.5em minus 0.4em\relax IEEE, 2013, pp.
  3304--3311.

\bibitem{eleftheriadis2015multi}
S.~Eleftheriadis, O.~Rudovic, and M.~Pantic, ``Multi-conditional latent
  variable model for joint facial action unit detection,'' in \emph{IEEE
  International Conference on Computer Vision}.\hskip 1em plus 0.5em minus
  0.4em\relax IEEE, 2015, pp. 3792--3800.

\bibitem{krahenbuhl2011efficient}
P.~Kr{\"a}henb{\"u}hl and V.~Koltun, ``Efficient inference in fully connected
  crfs with gaussian edge potentials,'' in \emph{Advances in Neural Information
  Processing Systems}, 2011, pp. 109--117.

\bibitem{zheng2015conditional}
S.~Zheng, S.~Jayasumana, B.~Romera-Paredes, V.~Vineet, Z.~Su, D.~Du, C.~Huang,
  and P.~H. Torr, ``Conditional random fields as recurrent neural networks,''
  in \emph{IEEE International Conference on Computer Vision}.\hskip 1em plus
  0.5em minus 0.4em\relax IEEE, 2015, pp. 1529--1537.

\bibitem{zhang2018identity}
Z.~Zhang, S.~Zhai, and L.~Yin, ``Identity-based adversarial training of deep
  cnns for facial action unit recognition,'' in \emph{British Machine Vision
  Conference}.\hskip 1em plus 0.5em minus 0.4em\relax BMVA Press, 2018, p. 226.

\bibitem{sanchez2018joint}
E.~Sanchez, G.~Tzimiropoulos, and M.~Valstar, ``Joint action unit localisation
  and intensity estimation through heatmap regression,'' in \emph{British
  Machine Vision Conference}.\hskip 1em plus 0.5em minus 0.4em\relax BMVA
  Press, 2018, p. 233.

\bibitem{newell2016stacked}
A.~Newell, K.~Yang, and J.~Deng, ``Stacked hourglass networks for human pose
  estimation,'' in \emph{European Conference on Computer Vision}.\hskip 1em
  plus 0.5em minus 0.4em\relax Springer, 2016, pp. 483--499.

\bibitem{pearl1988probabilistic}
J.~Pearl, \emph{Probabilistic Reasoning in Intelligent Systems: Networks of
  Plausible Inference}.\hskip 1em plus 0.5em minus 0.4em\relax Morgan Kaufmann,
  1988.

\bibitem{zhu2014multiple}
Y.~Zhu, S.~Wang, L.~Yue, and Q.~Ji, ``Multiple-facial action unit recognition
  by shared feature learning and semantic relation modeling,'' in
  \emph{International Conference on Pattern Recognition}.\hskip 1em plus 0.5em
  minus 0.4em\relax IEEE, 2014, pp. 1663--1668.

\bibitem{rumelhart1985learning}
D.~E. Rumelhart, G.~E. Hinton, and R.~J. Williams, ``Learning internal
  representations by error propagation,'' DTIC Document, Tech. Rep., 1985.

\bibitem{zhao2016joint}
K.~Zhao, W.-S. Chu, F.~De~la Torre, J.~F. Cohn, and H.~Zhang, ``Joint patch and
  multi-label learning for facial action unit and holistic expression
  recognition,'' \emph{IEEE Transactions on Image Processing}, vol.~25, no.~8,
  pp. 3931--3946, 2016.

\bibitem{corneanu2018deep}
C.~A. Corneanu, M.~Madadi, and S.~Escalera, ``Deep structure inference network
  for facial action unit recognition,'' in \emph{European Conference on
  Computer Vision}.\hskip 1em plus 0.5em minus 0.4em\relax Springer, 2018, pp.
  309--324.

\bibitem{simonyan2014very}
K.~Simonyan and A.~Zisserman, ``Very deep convolutional networks for
  large-scale image recognition,'' in \emph{International Conference on
  Learning Representations}, 2015.

\bibitem{he2016deep}
K.~He, X.~Zhang, S.~Ren, and J.~Sun, ``Deep residual learning for image
  recognition,'' in \emph{IEEE Conference on Computer Vision and Pattern
  Recognition}.\hskip 1em plus 0.5em minus 0.4em\relax IEEE, 2016, pp.
  770--778.

\bibitem{zhang2018interpretable}
Q.~Zhang, Y.~N. Wu, and S.-C. Zhu, ``Interpretable convolutional neural
  networks,'' in \emph{IEEE Conference on Computer Vision and Pattern
  Recognition}.\hskip 1em plus 0.5em minus 0.4em\relax IEEE, 2018, pp.
  8827--8836.

\bibitem{zhang2014bp4d}
X.~Zhang, L.~Yin, J.~F. Cohn, S.~Canavan, M.~Reale, A.~Horowitz, P.~Liu, and
  J.~M. Girard, ``Bp4d-spontaneous: a high-resolution spontaneous 3d dynamic
  facial expression database,'' \emph{Image and Vision Computing}, vol.~32,
  no.~10, pp. 692--706, 2014.

\bibitem{mavadati2013disfa}
S.~M. Mavadati, M.~H. Mahoor, K.~Bartlett, P.~Trinh, and J.~F. Cohn, ``Disfa: A
  spontaneous facial action intensity database,'' \emph{IEEE Transactions on
  Affective Computing}, vol.~4, no.~2, pp. 151--160, 2013.

\bibitem{valstar2015fera}
M.~F. Valstar, T.~Almaev, J.~M. Girard, G.~McKeown, M.~Mehu, L.~Yin, M.~Pantic,
  and J.~F. Cohn, ``Fera 2015-second facial expression recognition and analysis
  challenge,'' in \emph{IEEE International Conference and Workshops on
  Automatic Face and Gesture Recognition}, vol.~6.\hskip 1em plus 0.5em minus
  0.4em\relax IEEE, 2015, pp. 1--8.

\bibitem{zhang2016multimodal}
Z.~Zhang, J.~M. Girard, Y.~Wu, X.~Zhang, P.~Liu, U.~Ciftci, S.~Canavan,
  M.~Reale, A.~Horowitz, H.~Yang, J.~F. Cohn, Q.~Ji, and L.~Yin, ``Multimodal
  spontaneous emotion corpus for human behavior analysis,'' in \emph{IEEE
  Conference on Computer Vision and Pattern Recognition}.\hskip 1em plus 0.5em
  minus 0.4em\relax IEEE, 2016, pp. 3438--3446.

\bibitem{lucey2010extended}
P.~Lucey, J.~F. Cohn, T.~Kanade, J.~Saragih, Z.~Ambadar, and I.~Matthews, ``The
  extended cohn-kanade dataset (ck+): A complete dataset for action unit and
  emotion-specified expression,'' in \emph{IEEE Conference on Computer Vision
  and Pattern Recognition Workshops}.\hskip 1em plus 0.5em minus 0.4em\relax
  IEEE, 2010, pp. 94--101.

\bibitem{zhao2016deep}
K.~Zhao, W.-S. Chu, and H.~Zhang, ``Deep region and multi-label learning for
  facial action unit detection,'' in \emph{IEEE Conference on Computer Vision
  and Pattern Recognition}.\hskip 1em plus 0.5em minus 0.4em\relax IEEE, 2016,
  pp. 3391--3399.

\bibitem{walecki2017deep}
R.~Walecki, O.~Rudovic, V.~Pavlovic, B.~Schuller, and M.~Pantic, ``Deep
  structured learning for facial action unit intensity estimation,'' in
  \emph{IEEE Conference on Computer Vision and Pattern Recognition}.\hskip 1em
  plus 0.5em minus 0.4em\relax IEEE, 2017, pp. 5709--5718.

\bibitem{jia2014caffe}
Y.~Jia, E.~Shelhamer, J.~Donahue, S.~Karayev, J.~Long, R.~Girshick,
  S.~Guadarrama, and T.~Darrell, ``Caffe: Convolutional architecture for fast
  feature embedding,'' in \emph{ACM International Conference on
  Multimedia}.\hskip 1em plus 0.5em minus 0.4em\relax ACM, 2014, pp. 675--678.

\bibitem{sutskever2013importance}
I.~Sutskever, J.~Martens, G.~Dahl, and G.~Hinton, ``On the importance of
  initialization and momentum in deep learning,'' in \emph{International
  conference on machine learning}, 2013, pp. 1139--1147.

\bibitem{potts1952some}
R.~B. Potts, ``Some generalized order-disorder transformations,'' in
  \emph{Mathematical proceedings of the cambridge philosophical society},
  vol.~48, no.~1.\hskip 1em plus 0.5em minus 0.4em\relax Cambridge University
  Press, 1952, pp. 106--109.

\bibitem{shrout1979intraclass}
P.~E. Shrout and J.~L. Fleiss, ``Intraclass correlations: uses in assessing
  rater reliability,'' \emph{Psychological Bulletin}, vol.~86, no.~2, p. 420,
  1979.

\bibitem{fan2008liblinear}
R.-E. Fan, K.-W. Chang, C.-J. Hsieh, X.-R. Wang, and C.-J. Lin, ``Liblinear: A
  library for large linear classification,'' \emph{Journal of Machine Learning
  Research}, vol.~9, no. Aug, pp. 1871--1874, 2008.

\bibitem{zhong2015learning}
L.~Zhong, Q.~Liu, P.~Yang, J.~Huang, and D.~N. Metaxas, ``Learning multiscale
  active facial patches for expression analysis,'' \emph{IEEE Transactions on
  Cybernetics}, vol.~45, no.~8, pp. 1499--1510, 2015.

\bibitem{chu2017learning}
W.-S. Chu, F.~De~la Torre, and J.~F. Cohn, ``Learning spatial and temporal cues
  for multi-label facial action unit detection,'' in \emph{IEEE International
  Conference on Automatic Face and Gesture Recognition}.\hskip 1em plus 0.5em
  minus 0.4em\relax IEEE, 2017, pp. 25--32.

\bibitem{sandbach2013markov}
G.~Sandbach, S.~Zafeiriou, and M.~Pantic, ``Markov random field structures for
  facial action unit intensity estimation,'' in \emph{IEEE International
  Conference on Computer Vision Workshops}.\hskip 1em plus 0.5em minus
  0.4em\relax IEEE, 2013, pp. 738--745.

\bibitem{kaltwang2015latent}
S.~Kaltwang, S.~Todorovic, and M.~Pantic, ``Latent trees for estimating
  intensity of facial action units,'' in \emph{IEEE Conference on Computer
  Vision and Pattern Recognition}.\hskip 1em plus 0.5em minus 0.4em\relax IEEE,
  2015, pp. 296--304.

\bibitem{niu2016ordinal}
Z.~Niu, M.~Zhou, L.~Wang, X.~Gao, and G.~Hua, ``Ordinal regression with
  multiple output cnn for age estimation,'' in \emph{IEEE Conference on
  Computer Vision and Pattern Recognition}.\hskip 1em plus 0.5em minus
  0.4em\relax IEEE, 2016, pp. 4920--4928.

\bibitem{walecki2017copula}
R.~Walecki, O.~Rudovic, V.~Pavlovic, and M.~Pantic, ``Copula ordinal regression
  framework for joint estimation of facial action unit intensity,'' \emph{IEEE
  Transactions on Affective Computing}, no.~1, pp. 1--1, 2017.

\bibitem{linh2017deepcoder}
D.~Linh~Tran, R.~Walecki, O.~(Oggi)~Rudovic, S.~Eleftheriadis, B.~Schuller, and
  M.~Pantic, ``Deepcoder: Semi-parametric variational autoencoders for
  automatic facial action coding,'' in \emph{IEEE International Conference on
  Computer Vision}.\hskip 1em plus 0.5em minus 0.4em\relax IEEE, 2017, pp.
  3190--3199.

\bibitem{valstar2017fera}
M.~F. Valstar, E.~S{\'a}nchez-Lozano, J.~F. Cohn, L.~A. Jeni, J.~M. Girard,
  Z.~Zhang, L.~Yin, and M.~Pantic, ``Fera 2017-addressing head pose in the
  third facial expression recognition and analysis challenge,'' in \emph{IEEE
  International Conference on Automatic Face and Gesture Recognition}.\hskip
  1em plus 0.5em minus 0.4em\relax IEEE, 2017, pp. 839--847.

\end{thebibliography}

%
\begin{IEEEbiography}[{\includegraphics[width=1in,height=1.25in,clip,keepaspectratio]{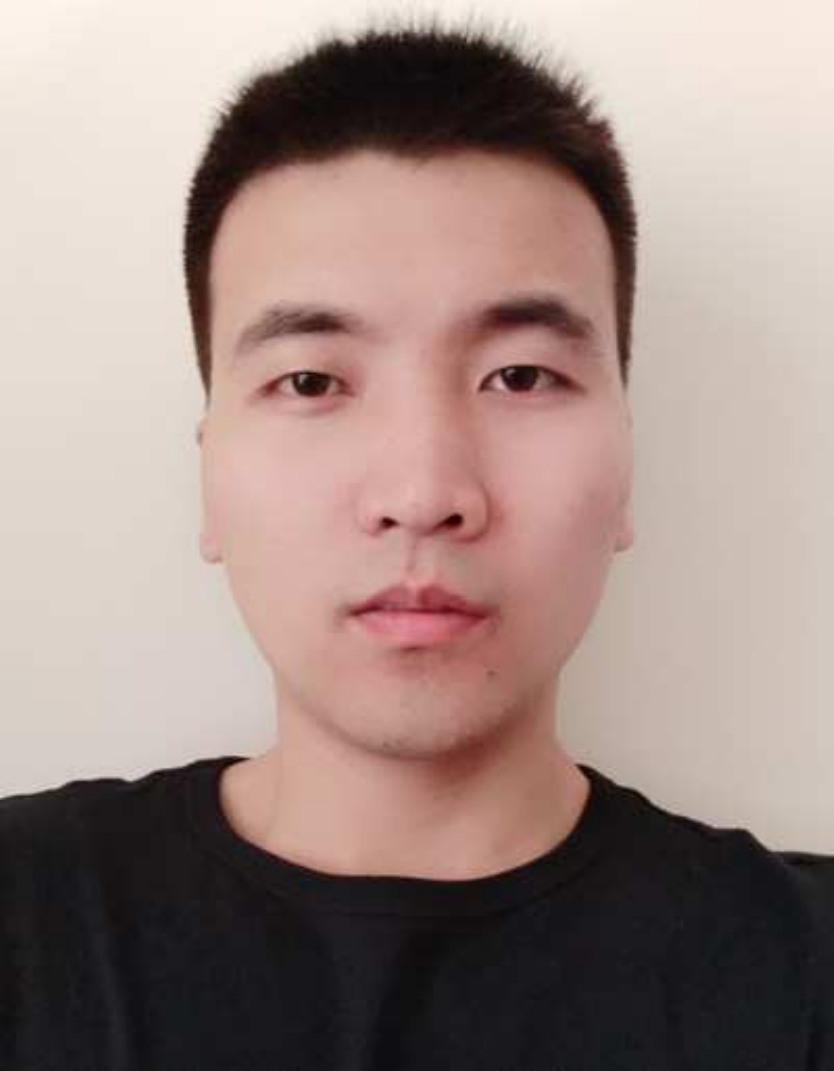}}]{Zhiwen Shao}
received his B.Eng. degree in Computer Science and Technology from the Northwestern Polytechnical University, China in 2015. He is now a Ph.D. candidate at the Department of Computer Science and Engineering, Shanghai Jiao Tong University, China. From 2017 to 2018, he was a joint Ph.D. student at the Multimedia and Interactive Computing Lab, Nanyang Technological University, Singapore. His research interests lie in face analysis and deep learning, in particular, facial expression recognition and face alignment.
\end{IEEEbiography}

\begin{IEEEbiography}[{\includegraphics[width=1in,height=1.25in,clip,keepaspectratio]{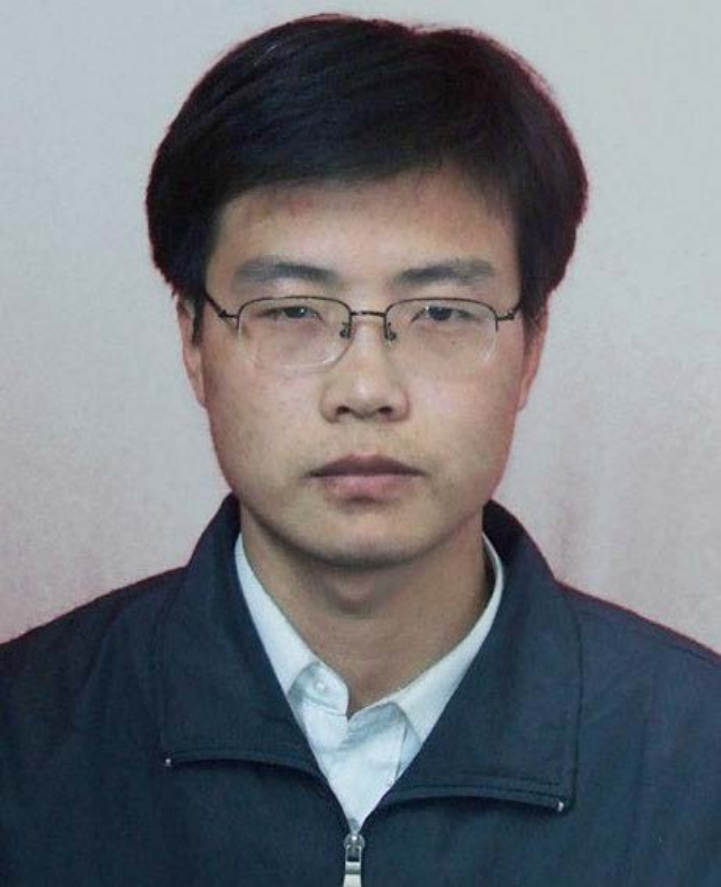}}]{Zhilei Liu}
received his Ph.D. degree in Computer Science from the University of Science and Technology of China in 2014. He is currently an Assistant Professor at the College of Intelligence and Computing, Tianjin University, China. From 2012 to 2013, he was a joint Ph.D. student at the Intelligent Systems Lab, Rensselaer Polytechnic Institute, USA. From 2017 to 2018, he was a Research Fellow at the Multimedia and Interactive Computing Lab, Nanyang Technological University, Singapore. His research interests cover multimedia computing, affective computing, machine learning and pattern recognition.
\end{IEEEbiography}

\begin{IEEEbiography}[{\includegraphics[width=1in,height=1.25in,clip,keepaspectratio]{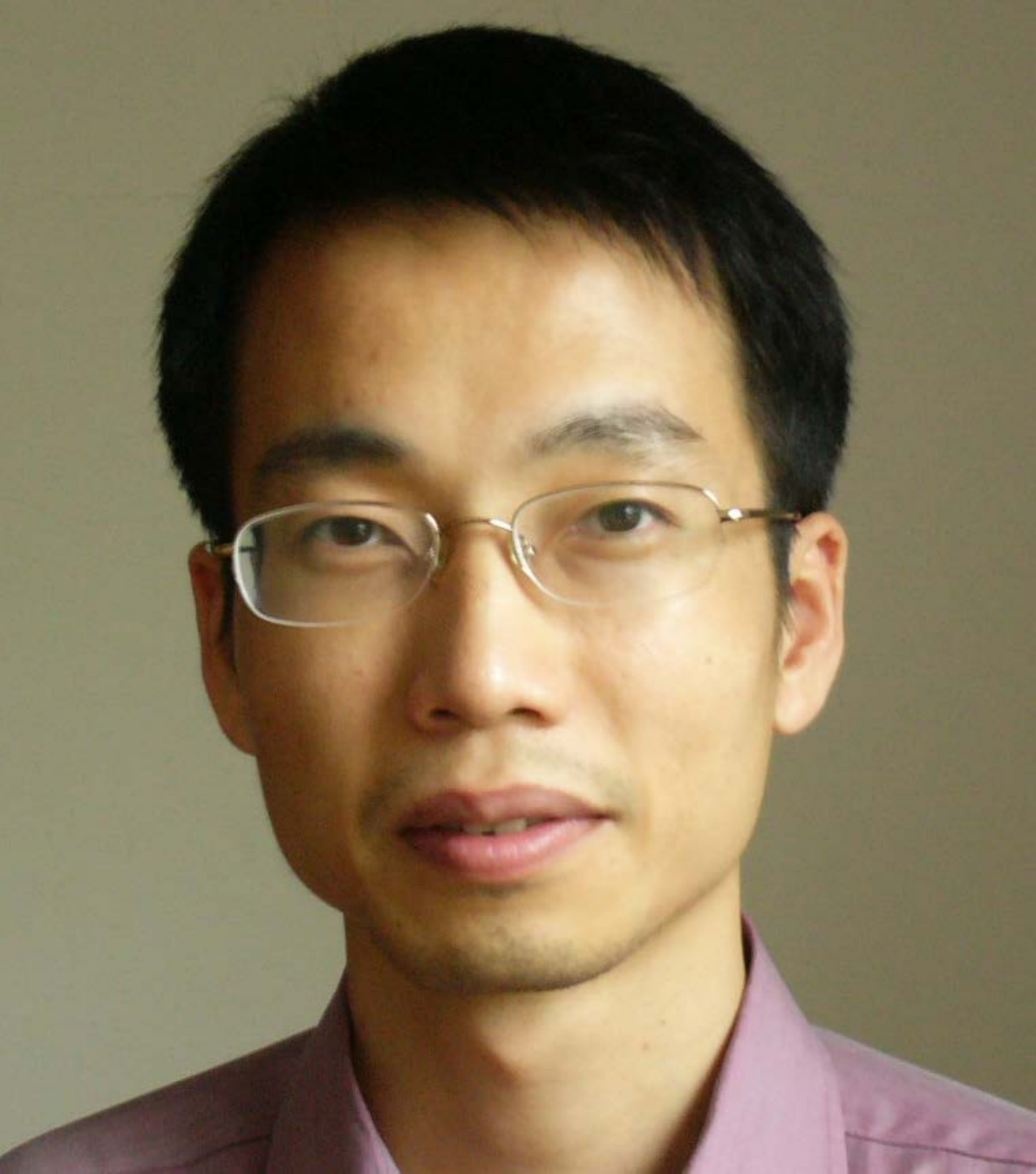}}]{Jianfei Cai}
received his Ph.D. degree from the University of Missouri-Columbia in 2002. He is currently a Full Professor and has served as the Head of Visual \& Interactive Computing Division and the Head of Computer Communication Division at the School of Computer Science and Engineering, Nanyang Technological University, Singapore. He has published over 200 technical papers in international journals and conferences. His major research interests include multimedia, computer vision and visual computing. He has served as the leading Technical Program Chair for ICME 2012. He is currently an Associate Editor for IEEE TMM, and has also served as an Associate Editor for IEEE TIP and TCSVT.
\end{IEEEbiography}

\begin{IEEEbiography}[{\includegraphics[width=1in,height=1.25in,clip,keepaspectratio]{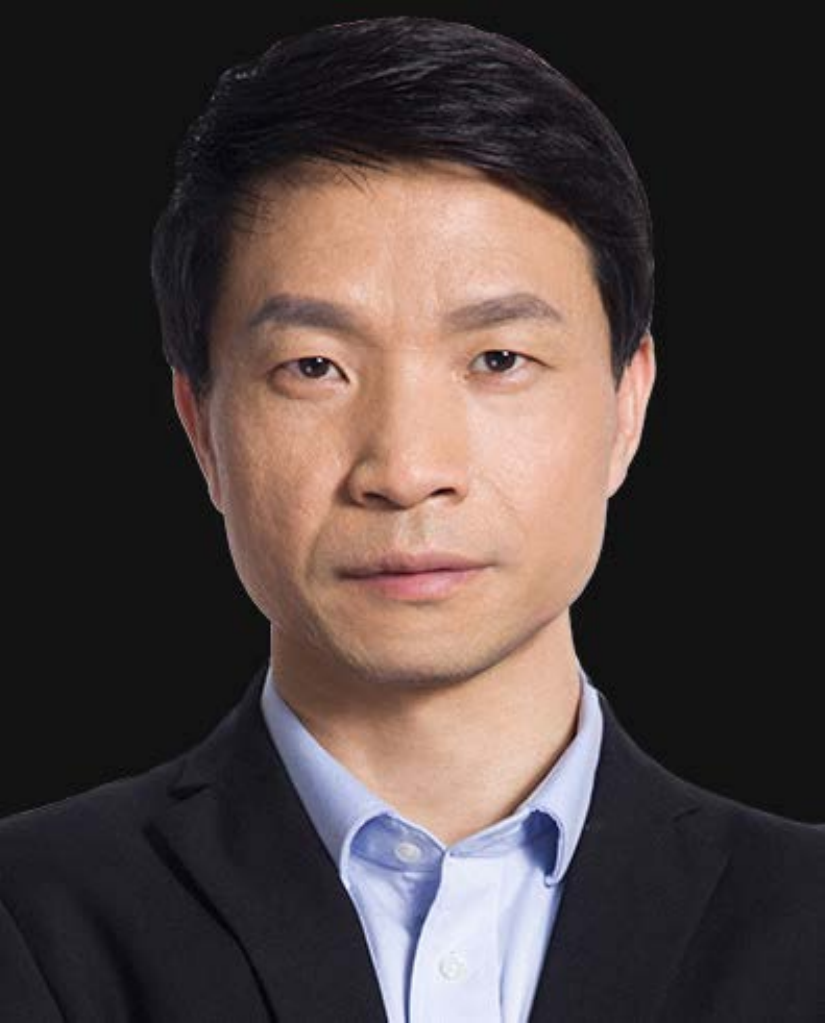}}]{Yunsheng Wu}
received his B.Eng. and M.S. degrees in Computer Science and Technology from the Peking University, China, in 2001 and 2004, respectively. He is now the General Manager of YouTu Lab at the Tencent Inc., China. Since joining Tencent in 2007, he has been responsible for many products such as QQ Video, QQ Tornado, Pitu, and Watermark Camera. His current research interests are image analysis and deep learning.
\end{IEEEbiography}

\begin{IEEEbiography}[{\includegraphics[width=1in,height=1.25in,clip,keepaspectratio]{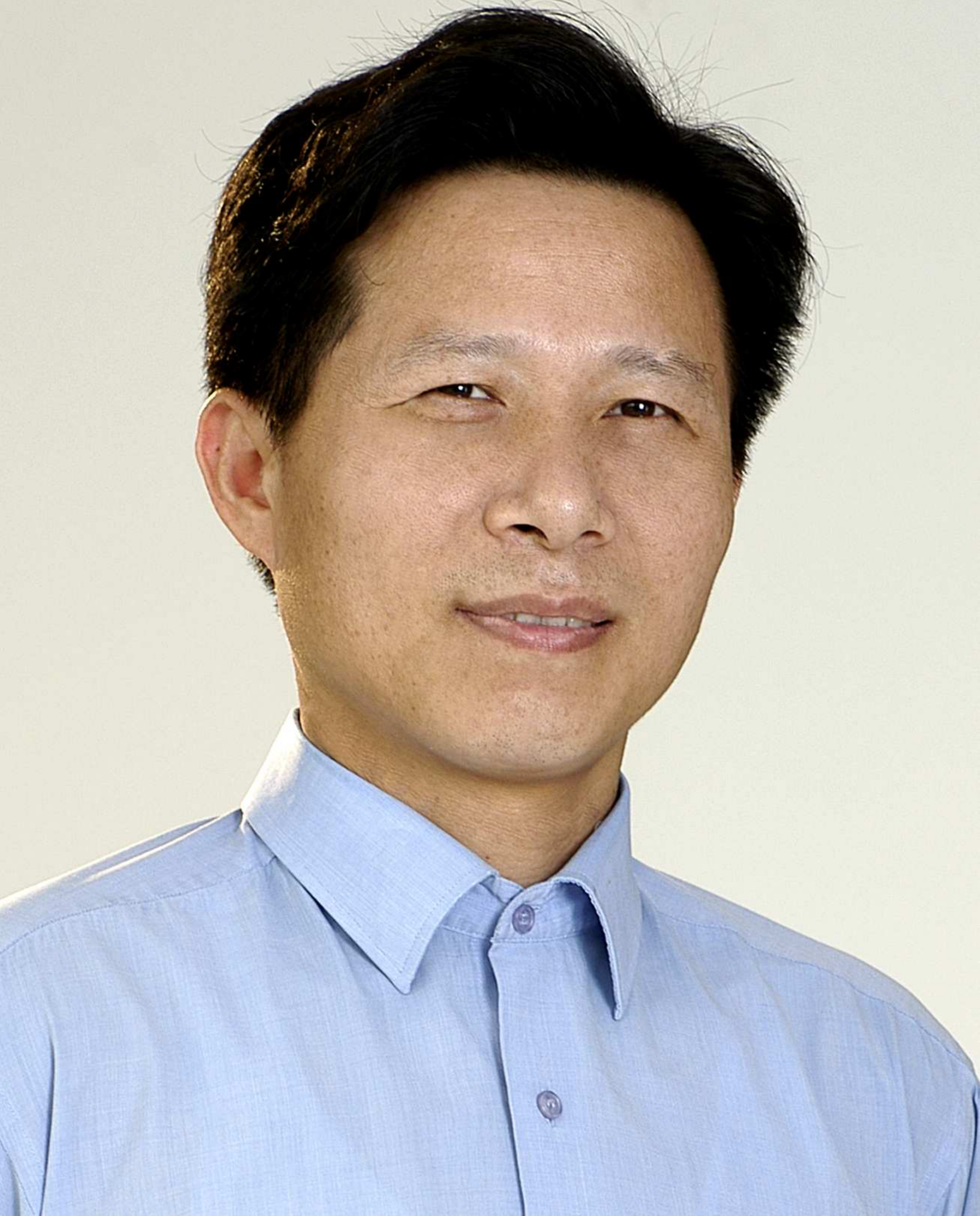}}]{Lizhuang Ma}
received his B.S. and Ph.D. degrees from the Zhejiang University, China in 1985 and 1991, respectively. He is now a Distinguished Professor, Ph.D. Tutor, and the Head of the Digital Media Technology and Data Reconstruction Laboratory at the Department of Computer Science and Engineering, Shanghai Jiao Tong University, China. He was a Visiting Professor at the Frounhofer IGD, Darmstadt, Germany in 1998, and was a Visiting Professor at the Center for Advanced Media Technology, Nanyang Technological University, Singapore from 1999 to 2000. He has published more than 200 academic research papers in both domestic and international journals. His research interests include computer aided geometric design, computer graphics, scientific data visualization, computer animation, digital media technology, and theory and applications for computer graphics, CAD/CAM.
\end{IEEEbiography}


%
%
%




\end{document}